\documentclass[10pt,twocolumn,letterpaper]{article}
\pdfoutput=1

\usepackage{cvpr}
\usepackage{times}
\usepackage{epsfig}
\usepackage{graphicx}
\usepackage{framed}
\usepackage{caption}
\usepackage{subcaption}
\usepackage{bbold}
\usepackage{adjustbox}
\usepackage{amsmath}
\usepackage{amssymb}
\usepackage{epstopdf}
\usepackage{amsthm}
\usepackage{mathtools}

\usepackage[pagebackref=true,breaklinks=true,letterpaper=true,colorlinks,bookmarks=false]{hyperref}

\cvprfinalcopy 

\newtheorem{theorem}{Theorem}

\DeclareMathOperator{\E}{\mathbf{E}}
\makeatletter
\let\@fnsymbol\@arabic

\def\cvprPaperID{1694} 

\pdfoutput=1
\ifcvprfinal\fi
\begin{document}

\newcommand{\vncomment}[1]{\textcolor{blue}{\bf \small #1 --VN}}
\newcommand{\srcomment}[1]{\textcolor{red}{\bf \small #1 --SR}}
\newcommand*\samethanks[1][\value{footnote}]{\footnotemark[#1]}

\title{Active learning with version spaces for object detection \thanks{An interested reader can find the proofs of theorems, details of the query function and more results at https://arxiv.org/pdf/1611.07285v1.pdf}}

\author{Soumya Roy \thanks{Indian Institute of Technology, Kanpur}
\and
Vinay P. Namboodiri \samethanks[2]
\and
Arijit Biswas \thanks{Amazon Core Machine Learning}
}

\maketitle
\begin{abstract}
Given an image, we would like to learn to detect objects belonging to particular object categories. Common object detection methods train on large annotated datasets which are annotated in terms of bounding boxes that contain the object of interest. Previous works on object detection model the problem as a structured regression problem that ranks the correct bounding boxes more than the background ones. In this paper we develop algorithms which actively obtain annotations from human annotators for a small set of images, instead of all images, thereby reducing the annotation effort. Towards this goal, we make the following contributions:
\newline 1. We develop a principled version space based active learning method that solves for object detection as a structured prediction problem in a weakly supervised setting
\newline 2. We also propose two variants of the margin sampling strategy
\newline 3. We analyse the results on standard object detection benchmarks that show that with only 20\% of the data we can obtain more than 95\% of the localization accuracy of full supervision. Our methods outperform random sampling and the classical uncertainty-based active learning algorithms like entropy
\end{abstract}

\section{Introduction}
In object detection, we aim to accurately obtain bounding boxes in an image that contain objects of a particular category like bicycle, person or motorbike. We follow a common practice \cite{blaschko2008}\cite{girshick14CVPR} of proposal generation, feature extraction and classification to detect objects. However, to train for object detection we need accurate ground truth bounding boxes (annotations) of a large number of images. Annotating images is laborious and expensive. Moreover, not all training images are useful. We solve this problem using active learning for object detection such that only a few relevant images are chosen to be annotated. While active learning has been widely used for object classification \cite{Kapoor:2010:GPO:1747084.1747105}\cite{Long:2015:MMA:2919332.2920026}\cite{7552533}, it has not been as well explored for the problem of object detection where it is more useful. 

We model the object detection problem as a structured regression problem \cite{blaschko2008} that can be solved using a Structured Support Vector Machine (SSVM). A key insight in our approach is that we can view the task of structured regression as that of learning a binary classification model in a ``difference of feature'' space where the features are obtained from the different bounding boxes in each image. This allows us to develop an active learning algorithm using the principled version space approach of \cite{Tong:2002:SVM:944790.944793} in the ``difference of feature'' space for structured prediction. Version space based approach is a classical idea in machine learning introduced by Mitchell \cite{mitchell_version_space82}. This idea has been further used in active learning in \cite{Tong:2002:SVM:944790.944793}\cite{He:2004:MVS:1026711.1026715}\cite{a2ba47b632b84c3f9c744081cf281ff8}. The main idea in this approach is to consider a version space consistent with the training data seen so far. Then we consider those samples for labeling that maximally reduce the version space. But a version space based approach for a structured regression task such as object detection is not directly possible. We show that by using the ``difference of feature'' space representation, one can obtain a version space representation for object detection. Further, an advantage we obtain is version space based active learning has strong theoretical foundations unlike the classical methods like entropy based querying strategy. As we show through an experimental evaluation, they also perform better than comparable baseline active learning methods.

The main takeaway from this paper is a theoretically well founded active learning method for object detection that outperforms other comparable active learning techniques. Currently, these have been developed using Caffe features \cite{jia2014caffe} in a SSVM framework. They can also be suitably adapted to work jointly with deeply learned networks.

\section{Related Work}
Object classification and detection are widely explored topics in computer vision. It started with recognizing objects using simple geometry \cite{Mundy2006} to generating better features for object classification and detection using \cite{790410}\cite{Dalal05histogramsof}\cite{Bay2008346} to using better classifiers \cite{blaschko2008}\cite{5255236}\cite{vedaldi09multiple} to using Convolutional Neural Networks (CNN) like \cite{Krizhevsky_imagenetclassification}, R-CNN \cite{girshick14CVPR}, Fast R-CNN \cite{girshickICCV15fastrcnn}, Faster R-CNN \cite{ren15fasterrcnn} and SSD \cite{DBLP:journals/corr/LiuAESR15}. However, all these methods make use of full supervision for object detection in terms of explicit bounding box information.

Active learning \cite{Settles10activelearning} has been widely used for object classification \cite{4563068}\cite{5206627}\cite{Kapoor:2010:GPO:1747084.1747105}\cite{Vijayanarasimhan2011}\cite{Li:2013:AAL:2514950.2516203}\cite{Long:2015:MMA:2919332.2920026}\cite{7552533}\cite{YangMNCH15}, where given an image we want to label the presence or absence of an object of a particular class, and video annotation \cite{Lampert2009}\cite{vondrick2011}\cite{Vijayanarasimhan2012}\cite{Karasev_2014_CVPR}. However, this problem has not been much explored for object detection which requires structured prediction of the correct bounding box.
We discuss a few related works that have explored this problem. In \cite{5995430}, the authors propose a part-based detector for object detection amenable to SVM, and show how to identify its most uncertain instances using the simple margin of \cite{Tong:2002:SVM:944790.944793} in sub-linear time with a hashing-based solution. \cite{YGLG12} proposes an incremental learning approach for some fixed data collection,  like medical images, that continuously updates an object detector and detection thresholds as a user interactively corrects annotations proposed by the detector. Unlike \cite{YGLG12}, in the proposed approach, the annotation costs are assumed to be equal for every image.
Active learning should not be confused with active detection methods like \cite{chen14active} where the discriminative ability of an already trained base classifier is refined at test time using minimum human supervision.

Weakly supervised object detection has received considerable attention over the last few years \cite{Bilen14b} \cite{DBLP:journals/corr/SongGJMHD14}\cite{ DBLP:journals/corr/BilenV15}\cite{ DBLP:journals/corr/CinbisVS15}\cite{Wang2014}. In weakly supervised object detection, one aims to detect objects by using image level label information without using any ground truth bounding boxes. Our work is related to weakly supervised object detection, as we rely on image level labeling for positive images during training and active learning (weakly supervised setting). But our method is not limited to this setting. We adopt it as image level labeling is easily obtainable. We however, do not aim to do weakly supervised object detection in this paper. Rather, 
the querying strategies proposed in this paper can be used to select images for annotation from a pool of weakly supervised images in order to improve the supervised part of semi-supervised object detection algorithms.

Unlike heuristic approaches like entropy, version space approaches have a strong theoretical foundation \cite{Tong:2002:SVM:944790.944793}. Through this paper we provide, to the best of our knowledge, the first work that uses version spaces in the context of object detection.

\section{Approach}

The main technique that our method relies on is the version space reduction approach of \cite{Tong:2002:SVM:944790.944793}. The version space approach is a classical idea that relies on the duality between model hypothesis and feature space representation. Version space $V$ is the set of model hypotheses that are consistent or satisfy training data in terms of features. Hyperplanes in the feature space $F$ can be seen as points in the version space $V$ and points in $F$ become hyperplanes in $V$. This is because each feature point can be seen as constraining the set of hyperplanes that can correctly classify that point. Given this representation one can easily obtain algorithms to select features or samples to be labeled such that the version space (model space) is maximally reduced. However, in a structured regression problem such as object detection, this is not directly obtainable. For an image $x_i$ that contains an object of a particular category (e.g. bike) there would be a set of bounding boxes that would be positive and another set of bounding boxes that would be negative. A naive interpretation of this approach would be to create a separate instance for each bounding box with a positive or negative label. However, for each image, we would like to rank the positive bounding boxes higher than the negative ones. Thus we adopt an approach that relies on a ``difference of feature'' representation. Given this representation, we can obtain a version space based active learning method for this feature representation that solves for object detection. We propose five different version space based algorithms and compare these methods with several baselines proposed recently.
\begin{figure}
\centering 
 \includegraphics[width=0.5\textwidth]{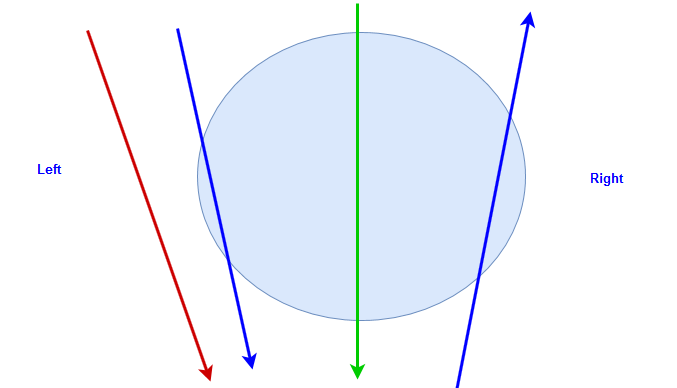}
 \caption{\textbf{Illustration of version space reduction}}
 \footnotesize{The circle represents the current version space and the arrows represent the different unlabeled data points. The solution space of an up arrow lies to the direction marked `Left' while that for a down arrow lies to the direction marked `Right'. The intersection of the current version space with the solution space of a data point is the version space obtained after adding that point to the training set. The red arrow does not intersect the version space and therefore has no effect while the green arrow maximally reduces the version space}
\label{fig:version_space}
\end{figure}
\subsection{Difference of Feature representation}
Consider a set of bounding boxes $Y_{x_i}$ from a set of bounding boxes $Y$ that are positive. All the bounding boxes $Y_{x_i}$ have an overlap greater than $0.5$ with the ground truth bounding box. Further consider a set $y_i$ that has the minimum overlap in $Y_{x_i}$. For any $m \in Y_{x_i}$ and $n \in y_i$, we obtain a ``difference of feature'' representation $F(x_i,m,n)=\psi(x_i,m)-\psi(x_i,n)$. The idea is that given this representation we can learn a model $w$ which ensures that all bounding boxes that are positive score positive when compared against the barely correct bounding boxes and score negative if they are not overlapping or have an overlap less than $0.5$ with the ground truth bounding boxes. The model is learned using the following formulation:

\begin{align} \label{eq:mssvm1}
\begin{split}
min_w \frac{\|w\|^2}{2} \nonumber\ \text{st}\ \forall i \forall y \in Y \backslash y_i \\ I(x_i,\ Y_{x_i},\ y)(w.\psi(x_i, y)- w.\psi(x_i, y_i)) \geq 1
\end{split}
\end{align}
Here $I$($x_i$, $Y_{x_i}$, $y$) = 1 if $y$ $\in$ $Y_{x_i}$ else -1. For a positive image, $y_i$ $\in$ $Y_{x_i}$ is the bounding box which has maximum loss (=50\%) with respect to the given annotations (the score decreases as the loss increases). For negative images, $y_i$ is the full image and $Y_{x_i}$ = \{\}, i.e. the null set. We term this formulation as Modified Structured Support Vector Machine formulation (M-SSVM) as this is similar to the Structured Support Vector Machine formulation for object detection \cite{blaschko2008}. This formulation is equivalent to a hard-margin SVM in the ``difference of feature'' space.

\begin{figure}
\centering 
 \includegraphics[width=0.4\textwidth]{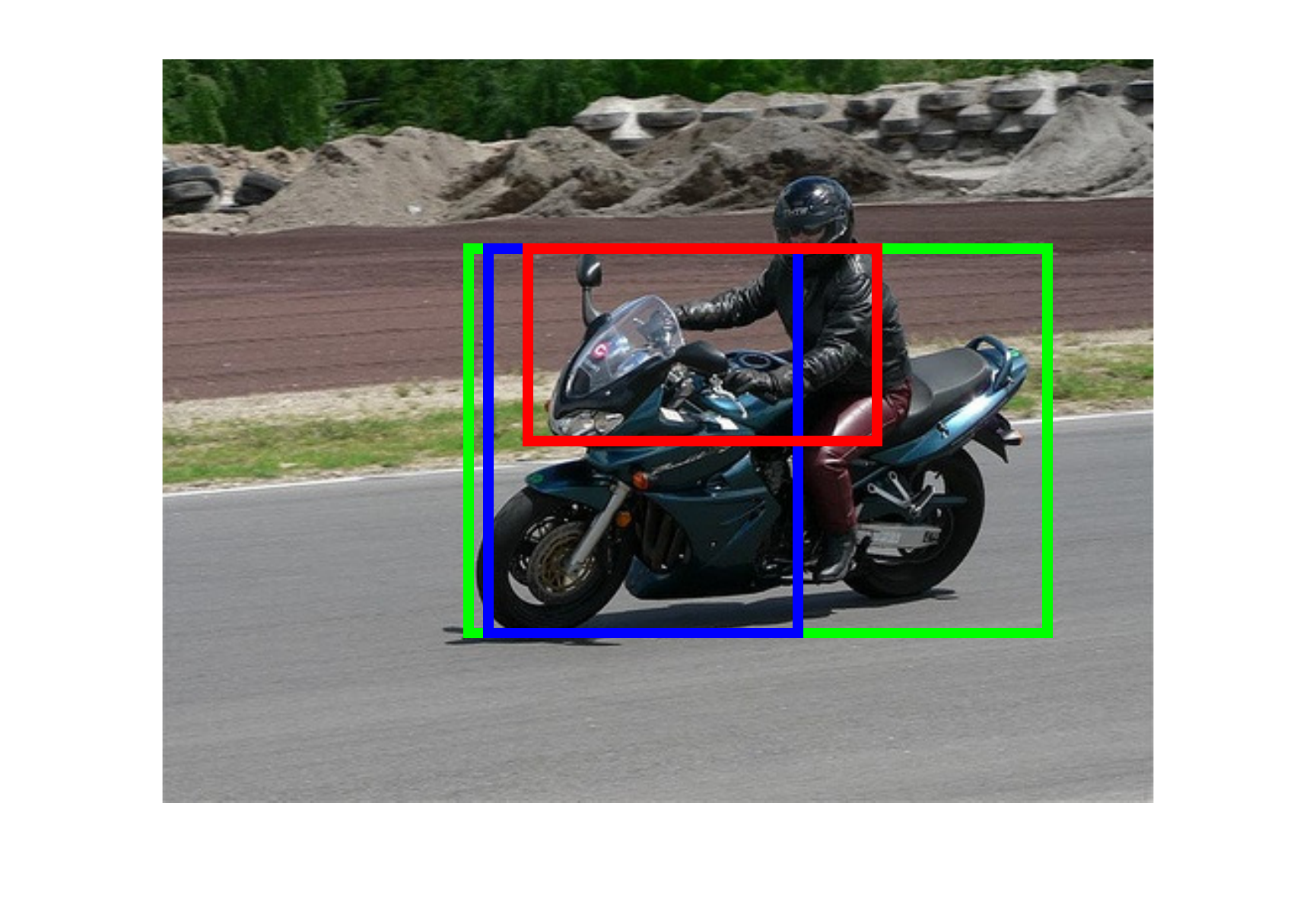}
 \caption{\textbf{Illustration of different feature representations}}
 \footnotesize{While the green bounding box is the ground truth bounding box, the blue box is a bounding box that has 50\% overlap with ground truth and the red box has lesser overlap}
\label{fig:bounding_box}
\end{figure}

\section{Method and its variants}
The active learning method relies on the version space representation in the ``difference of feature'' space. Our goal is to maximally reduce the version space while adding each annotated sample at every iteration. The area of the version space of an SVM is shown to be proportional to it's geometric margin in \cite{Tong:2002:SVM:944790.944793}. The version space $V$ is the set of hyperplanes in the parameter space $W$ that correctly classifies the currently labeled data points. If $V$ is current version space area and $V_{p_i}$ is the version space area after adding a newly labeled example $p_i$, then reduction in version space area $R_{p_i}$ = $V-V_{p_i}$. For maximum version space reduction, we choose that sample for user annotation, whose addition to the M-SSVM leads to minimum margin (inspired by \cite{He:2004:MVS:1026711.1026715}). However, in this procedure we need access to the ``future'' margin min$_{x_j \in Q}$min$_{y\in Y}$($|\hat{w}_{(t+1)}$.$F$($x_j$, $y_{j}$, $y|$) obtained after adding a newly annotated image $x_i$ to the training set $Q$, which already contains ($t$-1) annotated images. We estimate the ``future'' margin \footnote{A `*' is appended to any variable to indicate it's predicted value like $y^{*}_i$ is the predicted value of $y_i$ for image $x_i$} using an alternate prediction and training algorithm.

\subsection{Alternative Prediction-Training Algorithm (APT)}
\label{apt}
The  algorithm used for calculating ``future'' margin after adding unannotated image $x_i$ to the training set. is as follows:
\begin{enumerate}
\setlength\itemsep{0.2em}
\item Select the $y_j$ from the previously annotated images $x_j$ $\in$ $Q$ which has the minimum score with respect to $w_t$. Call it $divider$.
\item Sort ($x_i$, $y$) in descending order according to score $w_t$.$\psi(x_i,\ y)$.
\item Select $y \in Y$ that comes before $divider$ in the sorted list and include such $y$ in $Y^{*}_{x_i}$ (prediction step).
\item Select the minimum-scored $y$ in $Y^{*}_{x_i}$ as $y_{i}^{*}$.
\item Train the classifier to obtain $w^{*}_{(t+1)}$ (training step).
\\Continue steps 2 to 5 till convergence.
\item Recalculate $y^{*}_i$ from $w^{*}_{(t+1)}$.
\item Calculate ``future'' margins using $\hat{w}^{*}_{(t+1)}$, candidate windows of $x_i$ and $y^{*}_i$ as min$_{y\in Y}$($|\hat{w}^{*}_{(t+1)}$.$F$($x_i$, $y^{*}_{i}$, $y$)$|$).
\end{enumerate}

\begin{figure}
\centering 
 \includegraphics[width=0.5\textwidth]{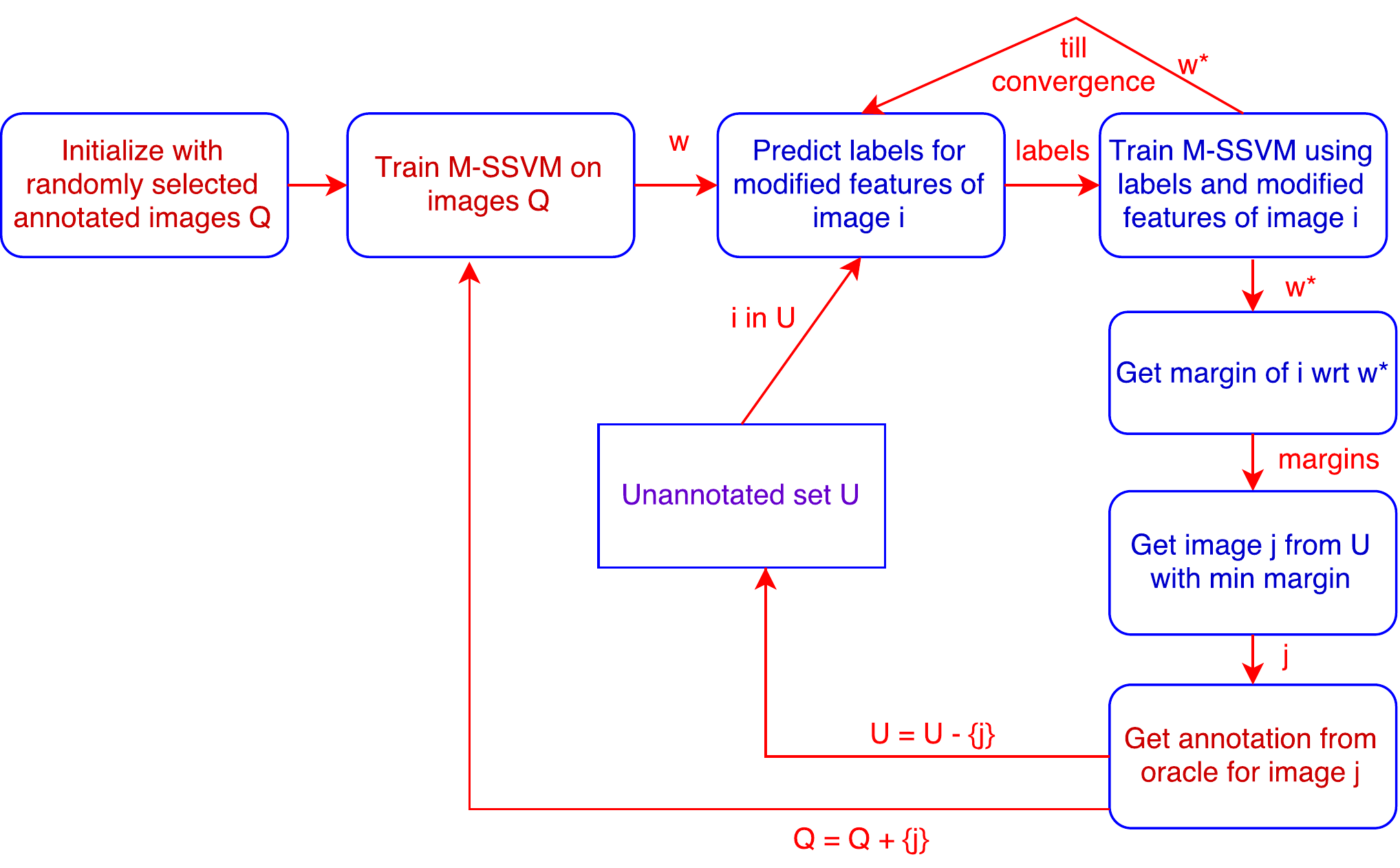}
 \caption{\textbf{Illustration of procedure for version space based active learning}}
\label{fig:caption}
\end{figure}

This algorithm is illustrated in the figure \ref{fig:caption}. 

While this is the main method, we also consider two other variations of this method and two variations of the margin-based querying strategy of \cite{a2ba47b632b84c3f9c744081cf281ff8}.

\subsection{Optimistic Pessimistic approach (OPT)}

In this approach, we use the $y_j$ obtained from previously annotated images $x_j$ to approximate y$_i$ of an unannotated image x$_i$. Subsequently, we predict the ``future'' margin after adding $x_i$ to the training set.

The steps for the procedure are as follows:
\begin{enumerate}
\setlength\itemsep{0.2em}
\item Find $y_{i}^{*}$ as
\begin{itemize}
\item {\bf Pessimistic method:} Select that $y_j$ from the previously annotated images $x_j$ which has the maximum score with respect to $w_t$ (a pessimistic estimate), as $y_{i}^{*}$.
\item {\bf Optimistic method:} Select that $y_j$ from the previously annotated images $x_j$ which has the minimum score with respect to $w_t$ (an optimistic estimate), as $y_{i}^{*}$.
\end{itemize}
 We begin with the pessimistic method when $w_t$ is not ``stable'' (reduces the number of false-positives in $Y^{*}_{x_i}$) and after querying a few images, we use the optimistic method.
\item Sort ($x_i$, $y$) in descending order according to score $w_t$.$\psi(x_i,\ y)$.
\item Select $y\in Y$ that come before $y_{i}^{*}$ in the sorted list and include such $y$ in $Y^{*}_{x_i}$.
\item Train the classifier to obtain $w^{*}_{(t+1)}$.
\\Continue steps 2 to 4 till convergence.
\item Calculate ``future'' margins using $\hat{w}^{*}_{(t+1)}$, candidate windows of $x_i$ and $y^{*}_i$ as min$_{y\in Y}$($|\hat{w}^{*}_{(t+1)}$.$F$($x_i$, $y^{*}_{i}$, $y$)$|$).
\end{enumerate}
Empirically, APT outperforms OPT as the $y^{*}_i$, obtained from the previously annotated images, and $y_i$ might have very different features.

\subsection{Maximum Model Change (MC)}
In active learning several methods are based on choosing samples that would induce a maximum change in the model. This approach is based on obtaining a maximum payoff for any sample that needs to be added.
An equivalent method can be obtained by considering the measure of model change between a model obtained using the APT method and the current model to choose samples that would result in a maximum model change. We analyse the connection between our method and the maximum model change approach in the theoretical section. Using Theorem \ref{theorem:th1} as a motivation, we use similarity measures between $w^{*}_{(t+1)}$, obtained using APT method (upto step 5), and $w_t$ to query unannotated images. In this paper, we use cosine similarity as $\frac{cos^{-1}|w^{*}_{(t+1)}.w_t|}{\|w^{*}_{(t+1)}\|\|w_t\|}$.

\subsection{Approximate Simple Margin (SM)}
\label{sm}
The margin of a Structured Support Vector Machine is  $min_i\hat{w}.F(x_i,\ y_i,\ y_{i}^{-})$ where $y_i$ is the correct class and $y_{i}^{-}$ is an incorrect class with the highest score. In  an earlier work \cite{a2ba47b632b84c3f9c744081cf281ff8}, the authors select a sample that has the minimum margin where margin is the difference between max-scored class and class with the second highest score. However, in practice, this poses a problem as the Selective Search procedure \cite{Uijlings13}, used to generate candidate windows, will hardly generate windows with complete overlap with an annotation. Such a strategy may find the difference of margins between two data points away from the hyperplane. This is also one of the reasons why M-SSVM has been introduced in this paper as the margin scaling pushes the modified features of the annotations away from the hyperplane. As the training progresses, the annotations become less important in determining the hyperplane. M-SSVM also allows multiple annotations.

We next propose a crude approximation to the margin of an unannotated $x_i$ and a candidate window $y$ as $\hat{w}.F(x_i,\ y_i,\ y)$ $<=$ $|\hat{w}.\psi(x_i,\ y_i)|$ + $|\hat{w}.\psi(x_i,\ y)|$ $<$ max$_{(x,y) \in {A_U}}|\hat{w}.\psi(x,\ y)|$ + $|\hat{w}.
\psi(x_i,\ y)|$ where $A_U$ is the set of correct classes (here annotations) of the corresponding unannotated samples. Querying using this upper bound is equivalent to querying using
$|\hat{w}.\psi(x_i,\ y)|$. Thus the querying strategy becomes min$_i$max$_{y \in Y \backslash y_i}|\hat{w}.\psi(x_i,\ y)|$. For each image $x_i$, we can use the candidate windows to determine max$_{y \in Y \backslash y_i}|\hat{w}.\psi(x_i,\  y)|$. However, this might not hold if there are more than one annotations in an image.

This query strategy chooses that image where our classifier is less confident about the most likely candidate for an annotation or where the most incorrect candidate window is closer to the correct ones. Such an intuition makes this query strategy independent of the nature of the candidate windows generation method and the number of annotations in an image.

\subsection{Modified Simple Margin (MSM)}
An approach based on simple margin has been proposed earlier \cite{Tong:2002:SVM:944790.944793}\cite{ a2ba47b632b84c3f9c744081cf281ff8}. 
Here we propose a method that slightly modifies the simple margin querying function and is sometimes more efficient.

For each unannotated image i,
\begin{enumerate}
\setlength\itemsep{0.2em}
\item Using the current weight vector w, select the maximum scored and second maximum scored candidate window. Call them $b_i$ and $c_i$ respectively.
\item For each annotated image, we select the annotation with the maximum score and add to set $A$. We use a similarity measure, like Gaussian kernel, to find that annotation from A which is most similar to $b_i$. Call it $a_i$.
\item Find margin m$_i$ as:
\[
\begin{cases}
\hat{w}.F(x_i,\ b_i,\ c_i)\ \text{if}\ \hat{w}.\psi(x_i,\ b_i)\ >=\ \hat{w}.\psi(x_i,\ a_i) \\
\hat{w}.F(x_i,\ a_i,\ b_i)\ \text{otherwise}
\end{cases}
\]\end{enumerate}
We find the unannotated image with the minimum margin.

\section{Theoretical Analysis of the method}
In this section we provide the main statements of the theorems that provide a strong theoretical foundation for our method. Theorem \ref{theorem:th2} explores the relationship between the actual weight vector and the predicted weight vector (step 5 of APT). This results in a bound on the predicted weight vector with respect to the current weight vector. Next we obtain a theoretical connection of our strategy with maximum model change and the simple margin of \cite{Tong:2002:SVM:944790.944793} through Theorem \ref{theorem:th1}. In both the theorems, we assume that SGD is the training procedure where, at step t, $\eta_t$ $>$ 0 is the learning rate and $\lambda$ $>=$ 0 is a regularization parameter used to control model complexity.

\subsection{Relation between $w_{t+1}$ and $w^{*}_{t+1}$}
\begin{theorem}
After ($t$ - 1) images are annotated, if $w_t$ is weight of the model, $p_i$ is the newly annotated image whose addition produces the model $w_{t+1}$, $m$ is the number of candidate windows for each image and $\theta_t$ is the classification accuracy of w$_t$ on the modified features of the candidate windows in p$_i$, then under reasonable assumptions:
\item $\|\E(w_{t+1}$-$w^{*}_{t+1})\|$$<=$(1-$\frac{\lambda}{mt}$)$\ln$(1+$\frac{1}{t-1}$)(2(1-$\theta_t$)+$\|y^{*}_{i}$-$y_{i}\|$)
\label{theorem:th2}
\end{theorem}

\subsection{Connection to other methods}
\begin{theorem}
At step $t$, if $w_t$ is weight of the model and ($p_i$, $q_i$)$_{i=1}^{n}$ are unlabeled examples, then under reasonable assumptions:
\\1. argmin$_{p_i}$ $w_{t+1}$.$w_t$ = argmin$_{p_i}$ $q_i$($w_{t+1}$.$p_i$)
\\2. argmin$_{p_i}\ q_i(w_t.p_i)$ $\implies$ argmin$_{p_i}$ $q_i.(\hat{w}_{t+1}.p_i)$
\label{theorem:th1}
\end{theorem}

\section{Baseline Methods}
We compare our method against comparable baselines based on other active learning strategies that have been used. While, the methods were not proposed for object detection, these are valid recently applied active learning approaches and they provide us with strong baselines for comparison. We define score s$_w$($x_i$,\ $y$) for image $x_i$, candidate window $y \in Y$ and model $w$ as $e^{\hat{w}.\psi(x_i,\ y)}$ where $Y$ is the set of windows obtained from the Selective Search method \cite{Uijlings13}. These querying strategies will be used on Structured Support Vector Machine (SSVM) as baselines for comparison. 
\subsection{Maximum Entropy (ENT)}
If $\mathbf{P}$($y|x_i$) =  $\frac{s_w(x_i,\ y)}{\sum_{y \in Y} s_w(x_i,\ y)}$ denotes the model's confidence about the candidate window $y$, the query strategy is max$_i$ $-\sum_{Y}\mathbf{P}$($y|x_i$)$\log \mathbf{P}$($y|x_i$). This is the AL-Entropy of \cite{wang2014new}.
\subsection{Min-Max (MM)}
A variation of the SM method, named min-max (MM), introduces a soft-max function in the querying strategy as min$_i$max$_{y \in Y}\frac{s_w(x_i,\ y)}{\sum_{y \in Y} s_w(x_i,\ y)}$. This is the AL-LC method of \cite{wang2014new}.
\subsection{Margin Sampling (MS)}
This is the margin-based query strategy of \cite{a2ba47b632b84c3f9c744081cf281ff8}. The querying strategy is to find that unannotated image which has the minimum margin where margin is determined by the difference between the highest scored and the second highest scored candidate windows. This is similar to the AL-MS method of \cite{wang2014new}. This is a popular measure for multi-class active learning and has been shown in \cite{wang2014new} to outperform other measures.
\section{Experimental Results}
\subsection{Setup}
We evaluate our methods on the PASCAL VOC2007 \cite{Pascal} and TU Graz-02 \cite{Opelt04} datasets. The ``Fast'' Selective Search method \cite{Uijlings13} is used to generate around 2000 candidate windows for each image. We use layer 7 of the Convolutional Neural Network (CNN) from Caffe \cite{jia2014caffe}, pre-trained on ILSVRC-2012 dataset, to extract a 4096 dimensional feature vector for each window.

The results for the active learning methods are reported on the test set using two measures Localization Accuracy and Localization Average Precision. Here, we restrict ourselves to localization and one object per test image. Localization Average Precision is the standard Average Precision (AP) measure when applied to only localization. Localization Accuracy is the percentage accuracy of localization (20\% is represented as 0.2) on the test set of the object, under question. This is similar to CorLoc \cite{Deselaers2012} but on the test set.

We train M-SSVM and SSVM using the faster but suboptimal Stochastic Gradient Descent (SGD) algorithm and the slower but optimal Cutting Plane (CP) method.
\subsubsection{PASCAL VOC2007}
The dataset has 2501 images in the training set, 2510 images in the validation set, 4952 images in the test set and 20 classes. The training and the validation set is merged. Like \cite{blaschko2008}, the training of the models is done only on the positive images of this merged training set but the annotations marked `difficult' are not left out during training.
\vspace{-0.2cm}

\subsubsection{TU Graz-02}
The dataset has 900 annotated images and 3 classes. The dataset for a given class is randomly and equally divided into a train and a test set.
\vspace{-0.2cm}

\subsection{Evaluation of Method and its variants}
\begin{figure}[ht]
\begin{framed}
\begin{subfigure}{0.37\textwidth}
\includegraphics[width=1.3\linewidth, height=2.9cm]{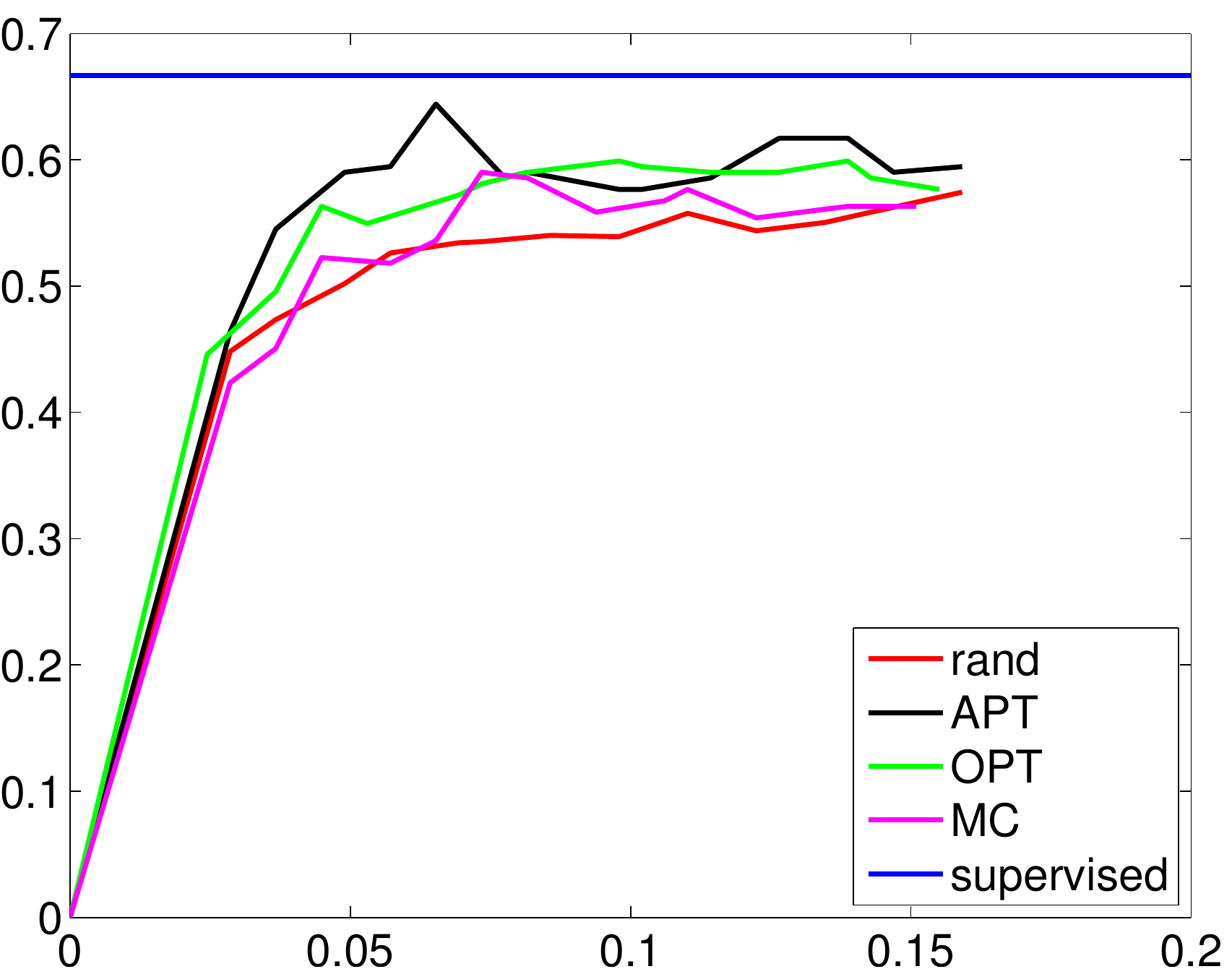}
\caption{motorbike}
\label{fig1a}
\end{subfigure}
\hspace{0.7cm}
\centering
\begin{subfigure}{0.37\textwidth}
\includegraphics[width=1.3\linewidth, height=2.9cm]{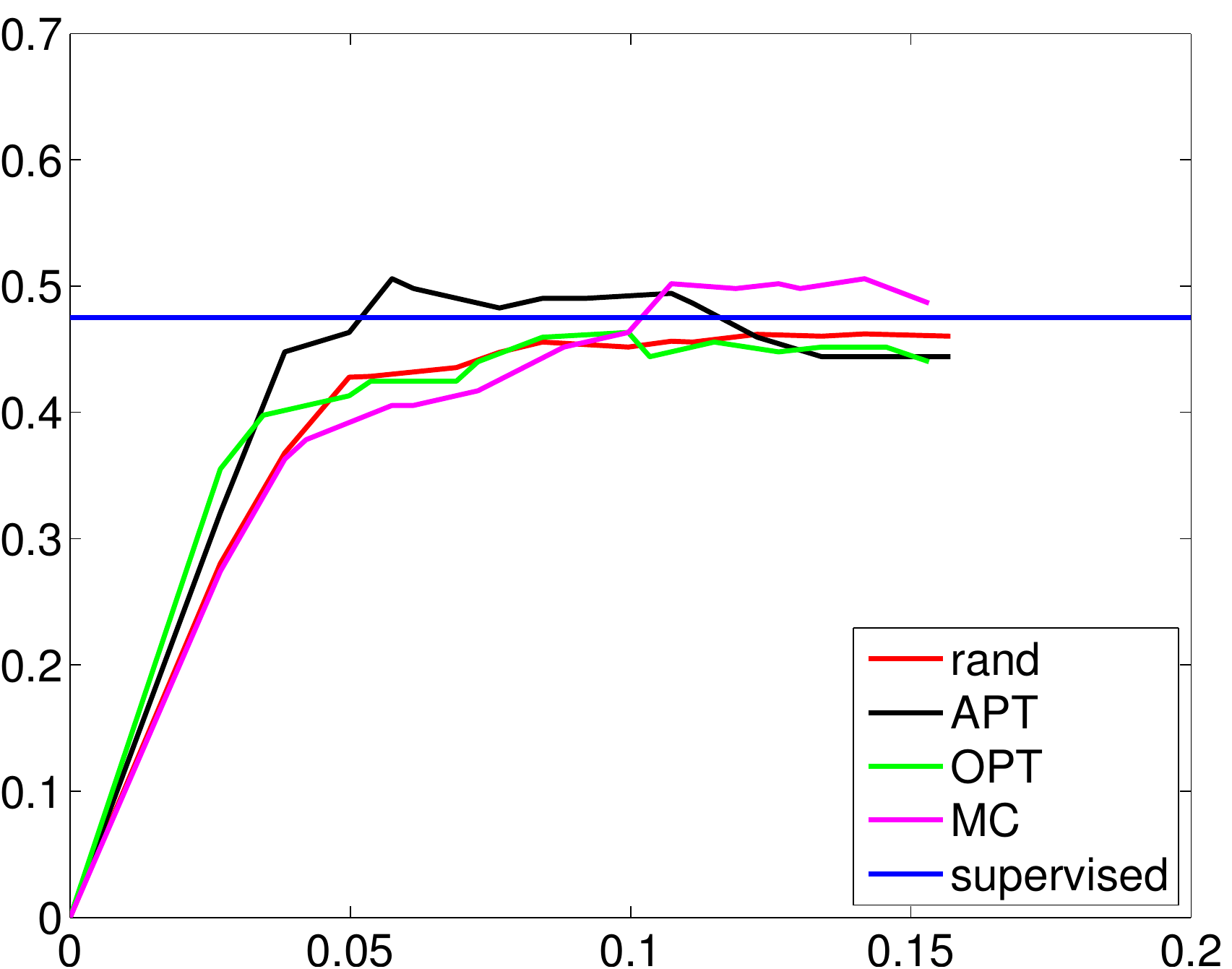}
\caption{train}
\label{fig1b}
\end{subfigure}
\centering
\begin{subfigure}{0.37\textwidth}
\includegraphics[width=1.3\linewidth, height=2.9cm]{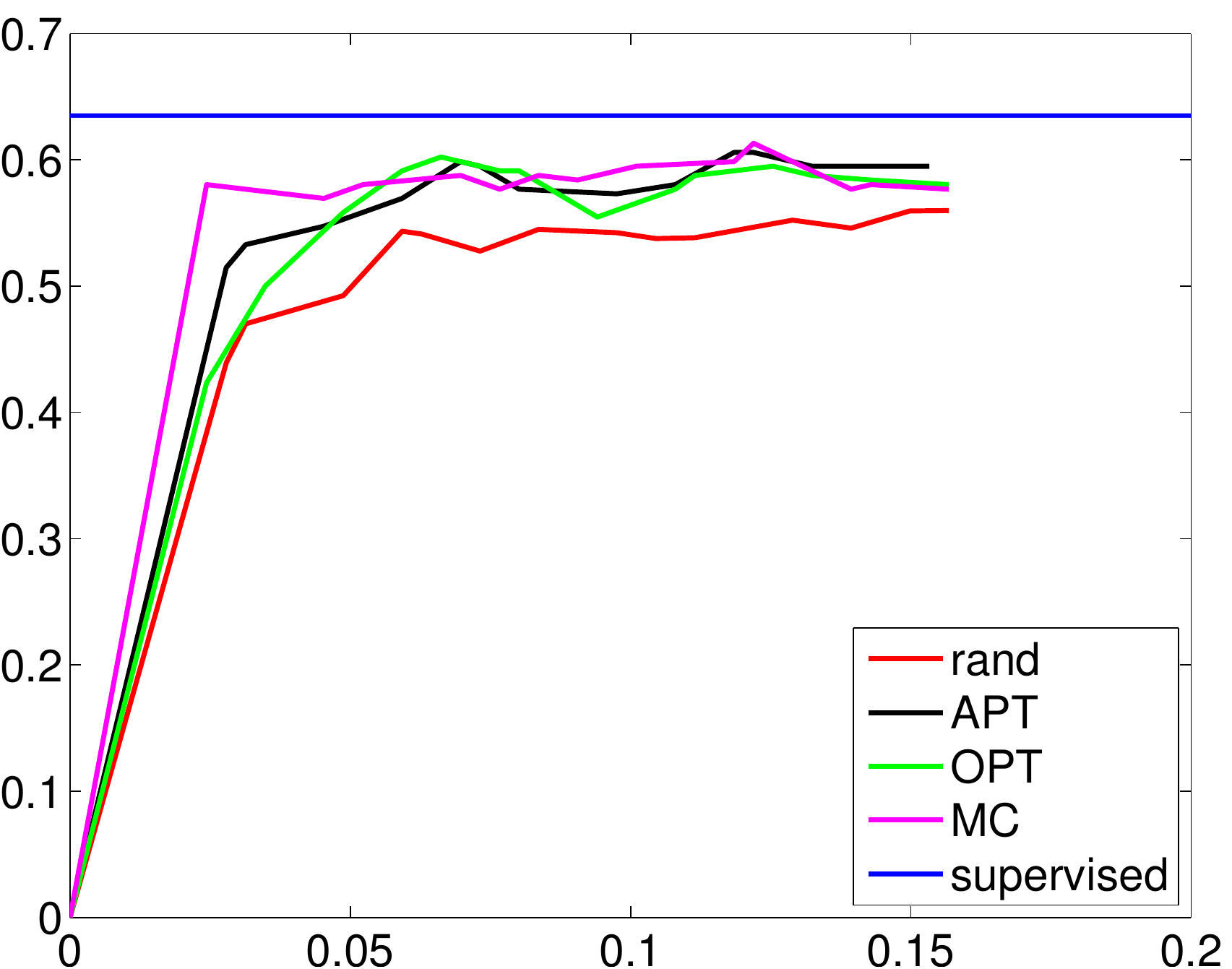}
\caption{horse}
\label{fig1c}
\end{subfigure}
\hspace{0.7cm}
\centering
\begin{subfigure}{0.37\textwidth}
\includegraphics[width=1.3\linewidth, height=2.9cm]{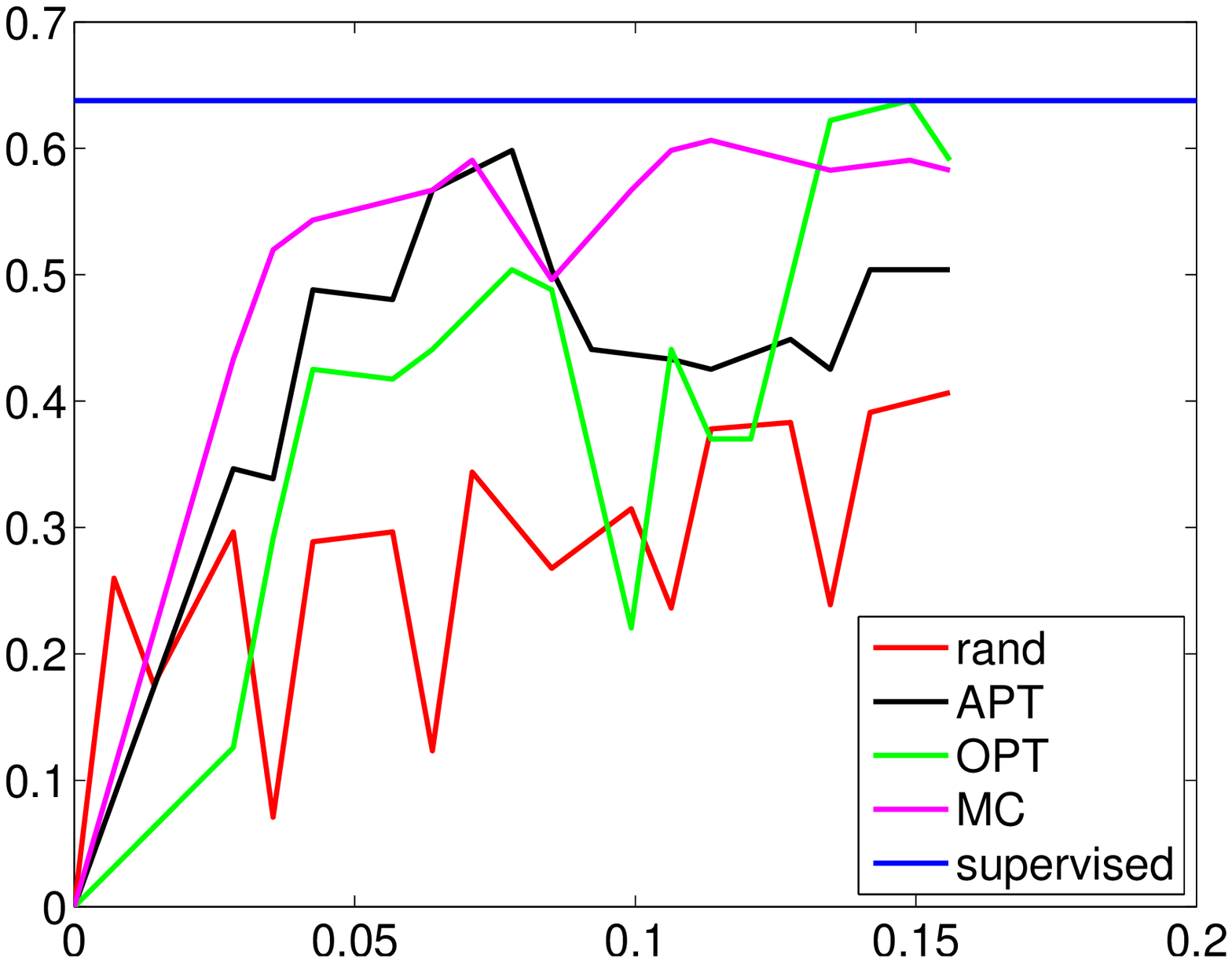}
\caption{cow}
\label{fig1d}
\end{subfigure}
\centering
\begin{subfigure}{0.37\textwidth}
\includegraphics[width=1.3\linewidth, height=2.9cm]{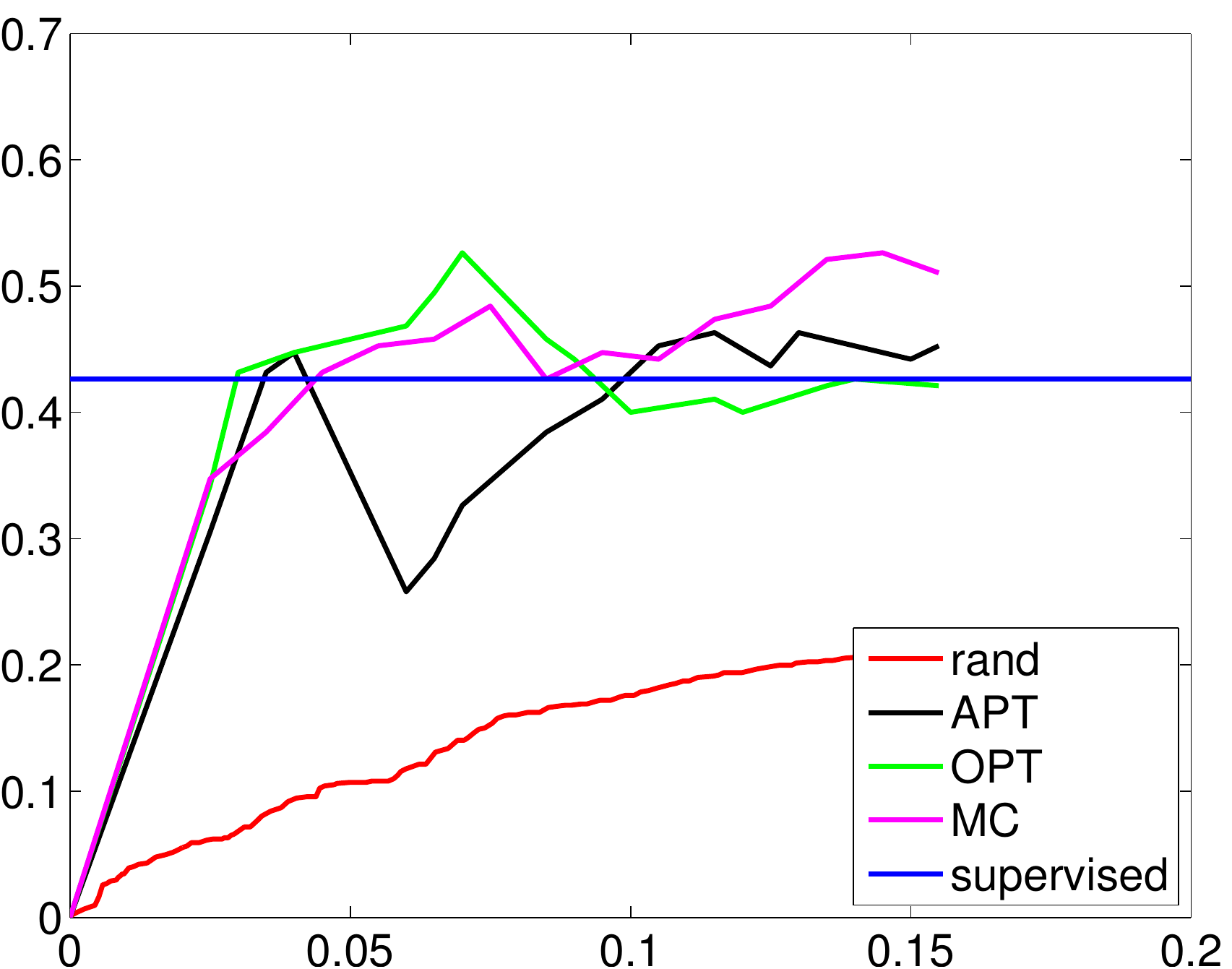}
\caption{dining table}
\label{fig1e}
\end{subfigure}
\hspace{0.7cm}
\centering
\begin{subfigure}{0.37\textwidth}
\includegraphics[width=1.3\linewidth, height=2.9cm]{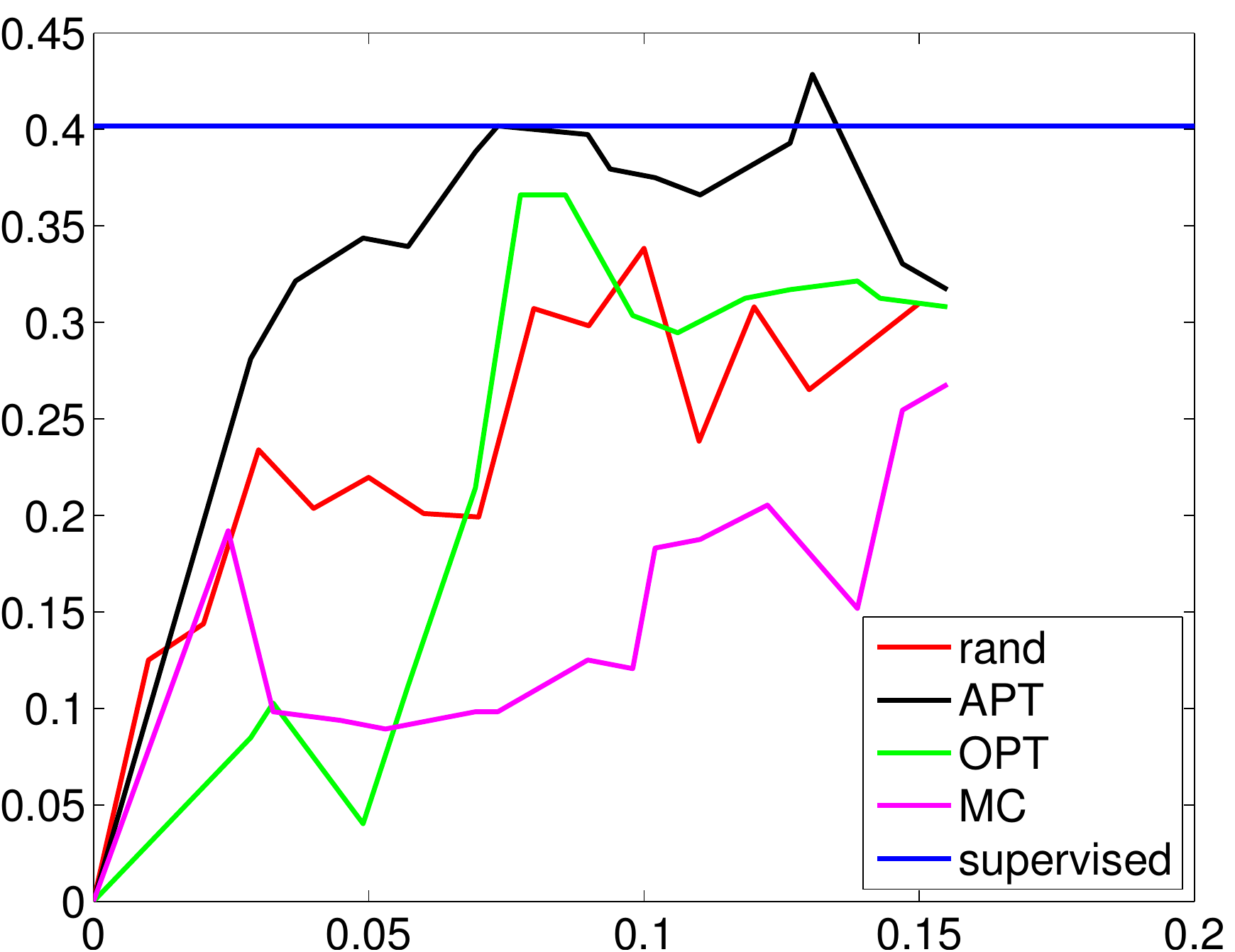}
\caption{potted plant}
\label{fig1f}
\end{subfigure}
\caption{\textbf{SGD version of the algorithm}}
\label{fig:sgd}
\footnotesize{Here the APT, OPT and MC methods are compared to random sampling. The optimization algorithm used for training is SGD. The graphs plot Localization Accuracy (y-axis) vs percentage of training images (x-axis). We use only 16\% of training images as we reach (or even cross) full supervision for most classes by that time.}
\end{framed}
\end{figure}

We train our M-SSVM on PASCAL VOC2007 using SGD, apply the APT, OPT and MC methods of active learning using SGD as the training method and report their corresponding Localization Accuracy. Then we train the hard-margin SSVM on the same dataset using SGD, run random sampling for 5 times and report the mean Localization Accuracy. The results for 6 diverse and interesting classes of PASCAL VOC2007 are given in Figure \ref{fig:sgd}. We choose the APT method for further experiments.
\vspace{-0.2cm}

\subsection{Comparison of Method with its baselines}
\label{others}
We train our M-SSVM using CP, apply APT method of active learning using SGD as the training method and report the results using Localization AP. For SSVM, we use the code \footnote{http://vision.princeton.edu/pvt/OnlineStructuralSVM} in \cite{Jia12} and set up the localization problem as shown in \cite{blaschko2008}. The same code is modified for M-SSVM. In addition, we perform active learning on SSVM using SM and MSM. We use the methods MM, ENT and MS on SSVM as baselines for comparison.
\subsubsection{PASCAL VOC2007} The results in Table \ref{tab:tab1} contain the maximum percentage of Localization AP, compared to that of full supervision reached by SSVM, obtained by APT, SM, MSM, ENT, MM and MS methods of active learning using 20\% samples. The mean Localization AP of full supervision is around 37.
\label{sec:results}
\begin{figure}[ht]
\begin{framed}
\centering
\resizebox{0.9\linewidth}{!}{%
 \begin{tabular}{*{7}{c}}
   class & \textit{APT} & \textit{SM} & \textit{MSM} & ENT & MM & MS\\
   \hline
   bike & \textbf{$>$100} & 95.6 & \textbf{$>$100} & 87.9 & 94.4 & 98.6\\
   train & \textbf{98.9} & 90.6 & 97.6 & 88.5 & 95.1 & 96.3\\
   cycle & \textbf{$>$100} & 80.5 & 85.8 & 81.4 & 81.9 & 98.9\\
   boat & \textbf{96.2} & 90.6 & 75.6 & 92.7 & \textbf{96.2} & 72.8\\
   bus & \textbf{97.8} & 77.6 & 77.3 & 76.5 & 94.6 & 97.3\\
   plane & \textbf{98.6} & 94.9 & 75.2 & 94.5 & 94.6 & 93.6\\
   car & \textbf{99.1} & 94.9 & 82.2 & 80 & 79 & 96.9\\
   horse & 95.9 & 97.5 & 79.7 & 79.5 & 98.3 & \textbf{99.9}\\
   dog & 91.9 & 90.9 & \textbf{94.1} & 91.8 & 89.3 & 91\\
   sheep & \textbf{$>$100} & 98.6 & 98.4 & 96.7 & \textbf{$>$100} & 92.1\\
   cow & \textbf{99} & 95.9 & 74.3 & 75 & 75.9 & 74.9\\
   cat & 98.4 & 74.2 & 84.6 & 82 & 80.8 & \textbf{$>$100}\\
   bird & 95.9 & 91.8 & 97.6 & 93.3 & 91.8 & \textbf{98.5}\\
   table & \textbf{$>$100} & 85.4 & 87.9 & 75.3 & 69.3 & 84\\
   sofa & \textbf{94} & 81 & 83.7 & 76.4 & 78.6 & 79.2\\
   plant & 94.9 & 98.2 & \textbf{$>$100} & 95.1 & 70.1 & 67.5\\
   tv & 86.8 & 87 & 87.9 & \textbf{88} & 86.9 & 87.3\\
   chair & 90 & 92.4 & \textbf{94.8} & 75.1 & 55 & 87.9\\
   bottle & 84 & 82.3 & \textbf{97} & 93.3 & 86.6 & 90.9\\
   person & 96.4 & 95.2 & 89.5 & 85 & 92.2 & \textbf{96.9}\\
   \hline
   mean & \textbf{95.9} & 89.8 & 88.2 & 85.4 & 85.5 & 90.2\\
   \hline
  \end{tabular}}
 \captionof{table}{\textbf{VOC2007 (as \% of full supervision)}}
 \label{tab:tab1}
 \end{framed}
\end{figure}
\subsubsection{TU Graz-02}  The results in Table \ref{tab:tab2} contain the maximum percentage of Localization AP, compared to that of full supervision reached by SSVM, obtained by APT, SM, MSM, ENT, MM and MS methods of active learning using 10\% samples. The mean Localization AP of full supervision is around 42. As this is an easier dataset than PASCAL VOC2007, we limit ourselves to 10\% of the training images.
 \begin{figure}[ht]
 \begin{framed}
\centering
  \resizebox{0.9\linewidth}{!}{%
 \begin{tabular}{*{7}{c}}
   class & \textit{APT} & \textit{SM} & \textit{MSM} & ENT & MM & MS\\
   \hline
   bike & \textbf{$>$100} & 80.3 & 86.2 & \textbf{$>$100} & 85.8 & 85\\
   person & 96.4 & \textbf{$>$100} & 93.7 & 87.2 & \textbf{$>$100} & 85.6\\
   cars & 99.6 & 98.6 & \textbf{$>$100} & 96.1 & \textbf{$>$100} & \textbf{$>$100}\\
   \hline
   mean & \textbf{98.7} & 93 & 93.3 & 94.4 & 95.3 & 90.2\\
   \hline
  \end{tabular}}
 \captionof{table}{\textbf{TU Graz-02 (as \% of full supervision)}}
 \label{tab:tab2}
 \end{framed}
\end{figure}
\vspace{-0.2cm}

\subsection{Discussion}
\label{discuss}
\vspace{-0.2cm}
\subsubsection{The curious case of random sampling}
Figure \ref{fig:sgd} shows the unreliability of random sampling. Random sampling works really well on `train', works moderately well on `motorbike' and `horse' and works poorly on `cow', `dining table' and `potted plant'.

Random sampling might choose images whose candidate windows do not intersect the current version space at all, thus yielding no improvement in the results. Moreover, in practice, annotating previously unannotated images might actually make the result worse than before because of the effect of outliers. This effect seems to become more pronounced for classes with small number of training images like `cow' and hence the wavy nature of their graphs. This effect is also the reason why some active learning methods cross the accuracy of full supervision.
\vspace{-0.2cm}
\subsubsection{How does APT fare?}
\vspace{-0.2cm}
Initially, due to low $\theta_t$, the baseline methods may perform better than APT (Theorem \ref{theorem:th2}). But the APT method performs consistently well for almost all the classes and mostly outperforms proposed alternatives like OPT and MC.

As seen from Table \ref{tab:tab1} and Table \ref{tab:tab2}, in classes with poor detection accuracy like `bottle', APT performs badly and in classes with better detection accuracy, like `motorbike', APT outperforms the other active learning methods. For example, classes like `motorbike' and `bicycle' cross full supervision while classes `chair' and `bottle' have (in terms of percentage) a mean accuracy of 87 for APT, mean accuracy of 87.4 for SM, mean accuracy of 95.9 for MSM, mean accuracy of 87.9 for ENT and mean accuracy of 89.4 for MS. In the end, the APT method, on average, achieves more than 95\% of full supervision with just 20\% of images.
\subsubsection{How do the baseline methods fare?}
\vspace{-0.2cm}
In spite of the practical difficulties, MS outperforms SM and MSM on PASCAL VOC2007. However, in classes with poor detection accuracy like `sofa', `potted plant', `chair' and `bottle', in terms of percentage mean accuracy is 93.9 for MSM while mean accuracy for MS is 81.4. ENT and MM are the worst performing methods on this dataset.

TU Graz-02 is an easier dataset and all the methods achieve more than 90\% of full supervision using only 10\% of training images. In this dataset, MM considerably outperforms MS.
\vspace{-0.2cm}
\subsubsection{Visual Examples}
\begin{figure*}[ht]
\begin{framed}
\centering
\includegraphics[width=.97\linewidth, height=9.5cm]{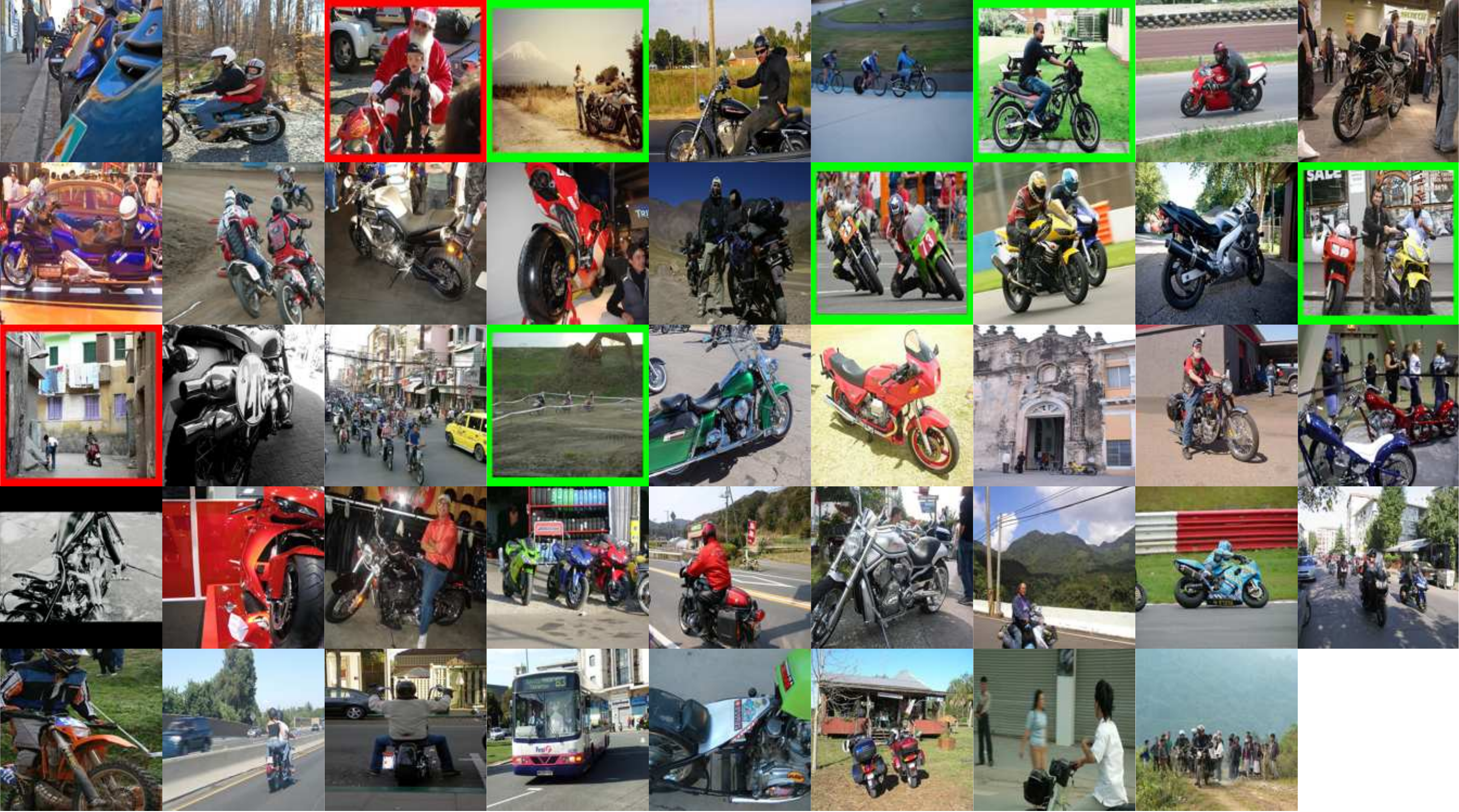}
\caption{\textbf{Images queried by APT}}
\label{fig:query_order}
\footnotesize{Here we show the images queried by each iteration of APT for the class `motorbike'. The images appear to be diverse. The algorithm prefers images containing multiple annotations (39\% of queried images vs 29\% of training images). The images with red border, when included in the training set, cause a significant drop in Localization Accuracy and can be termed `outliers'. On the other hand, the images with green border cause a significant rise in Localization Accuracy and can be called `influential'.}
\end{framed}
\end{figure*}

\begin{figure*}
\begin{framed}
\begin{subfigure}{0.2\textwidth}
\includegraphics[width=0.9\linewidth, height=2.7cm]{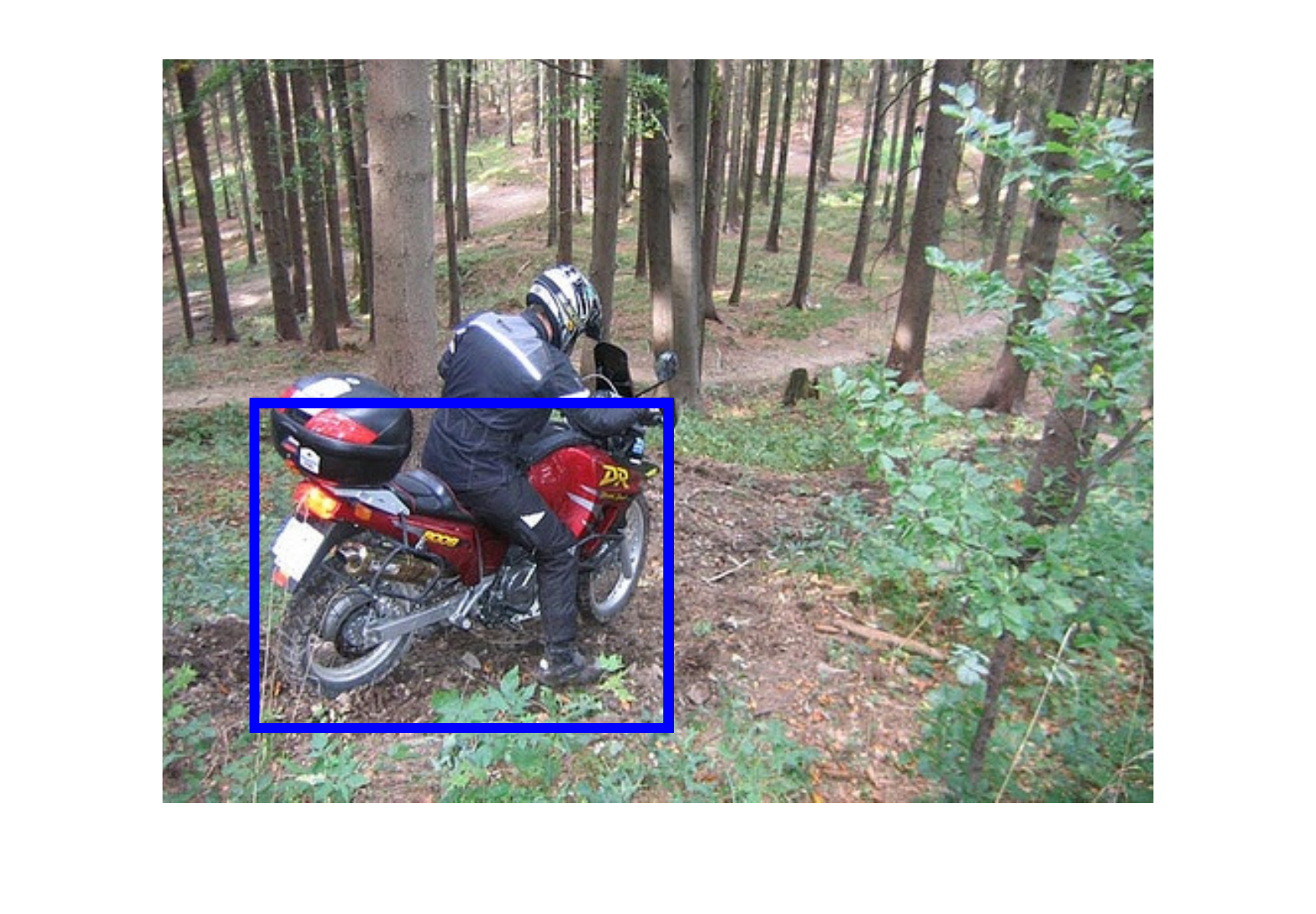}
\caption{bike}
\label{fig5a}
\end{subfigure}
\begin{subfigure}{0.2\textwidth}
\includegraphics[width=0.9\linewidth, height=2.7cm]{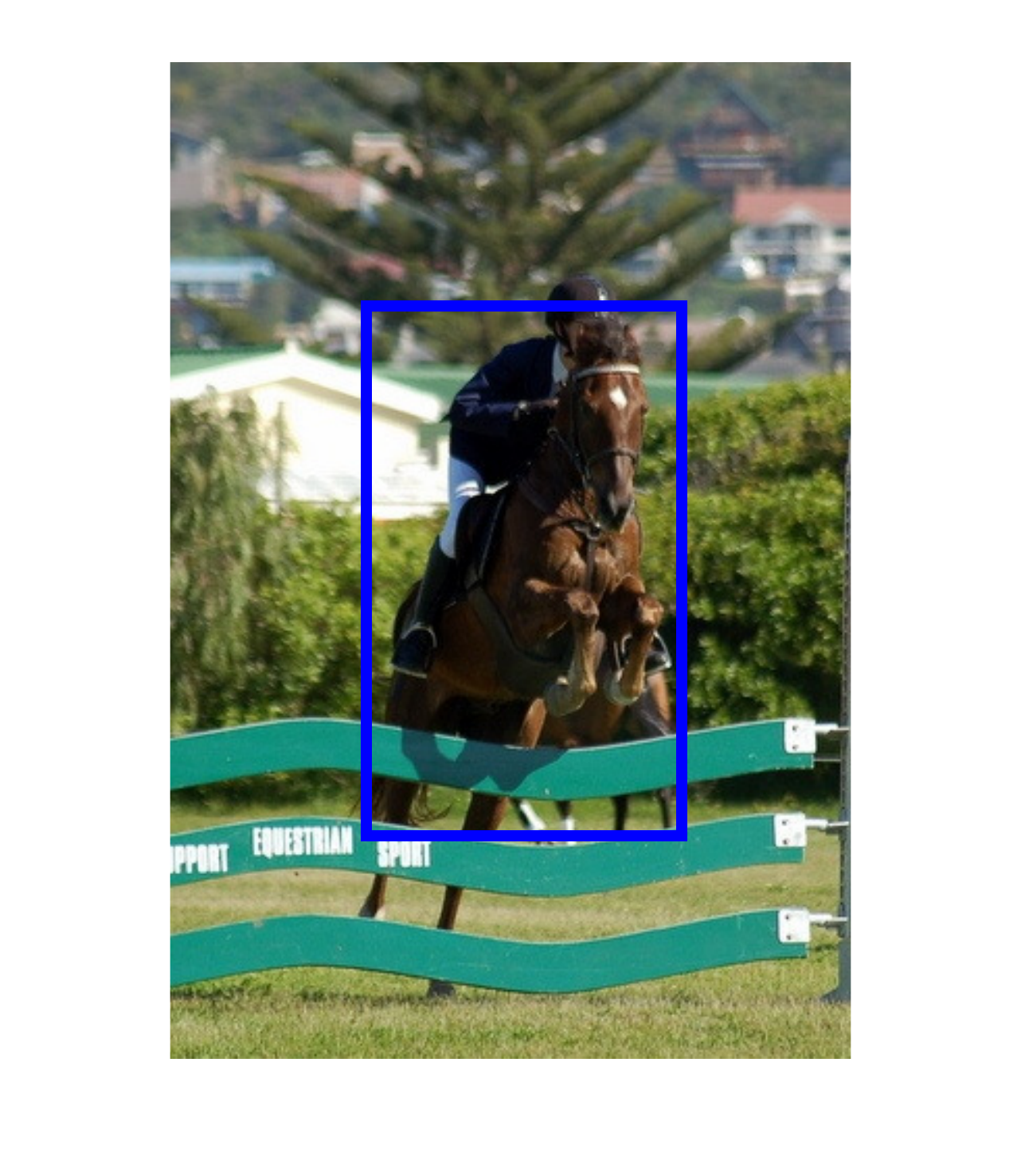}
\caption{horse}
\label{fig5b}
\end{subfigure}
\begin{subfigure}{0.2\textwidth}
\includegraphics[width=0.9\linewidth, height=2.7cm]{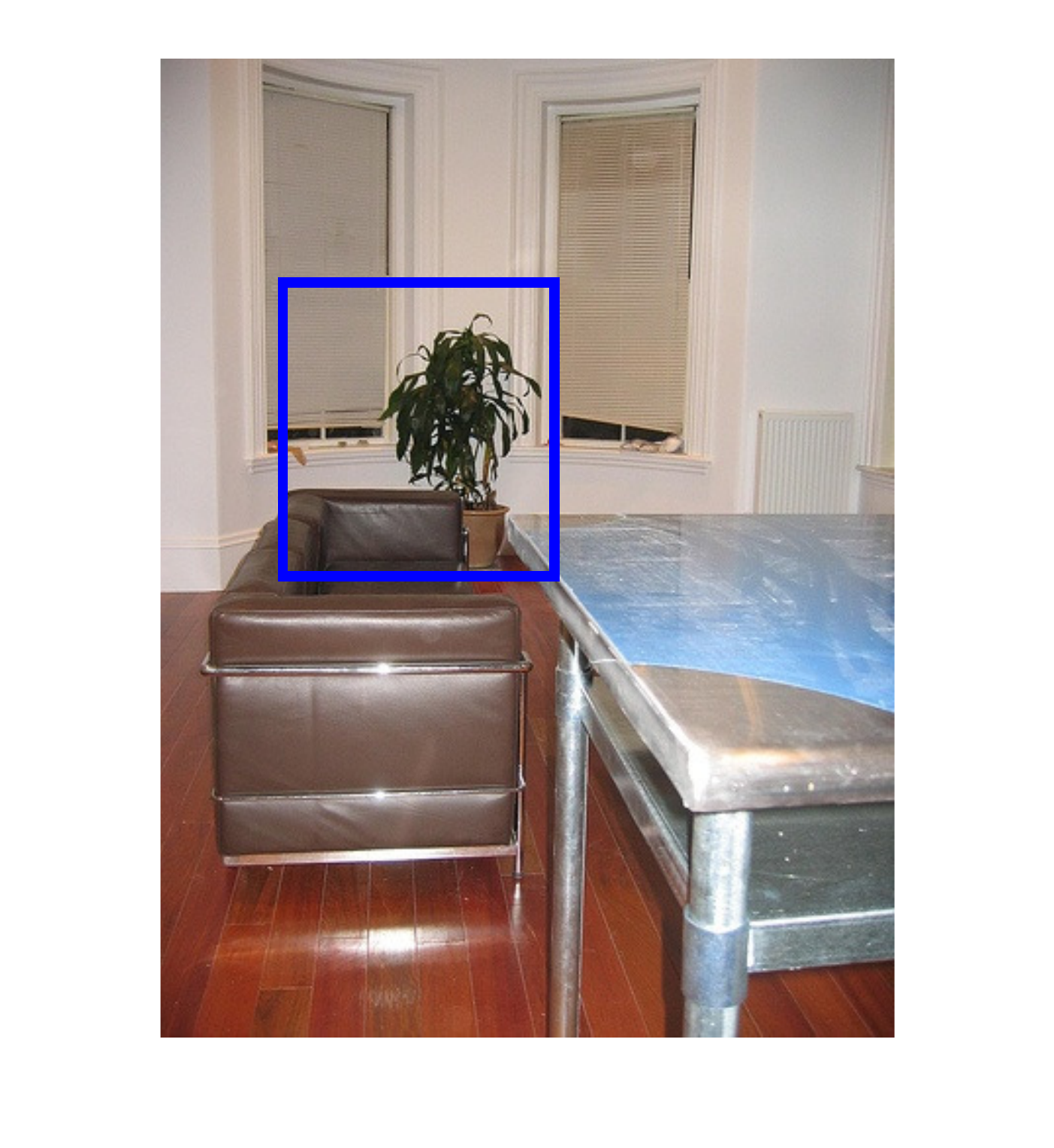}
\caption{plant}
\label{fig5c}
\end{subfigure}
\begin{subfigure}{0.2\textwidth}
\includegraphics[width=0.9\linewidth, height=2.7cm]{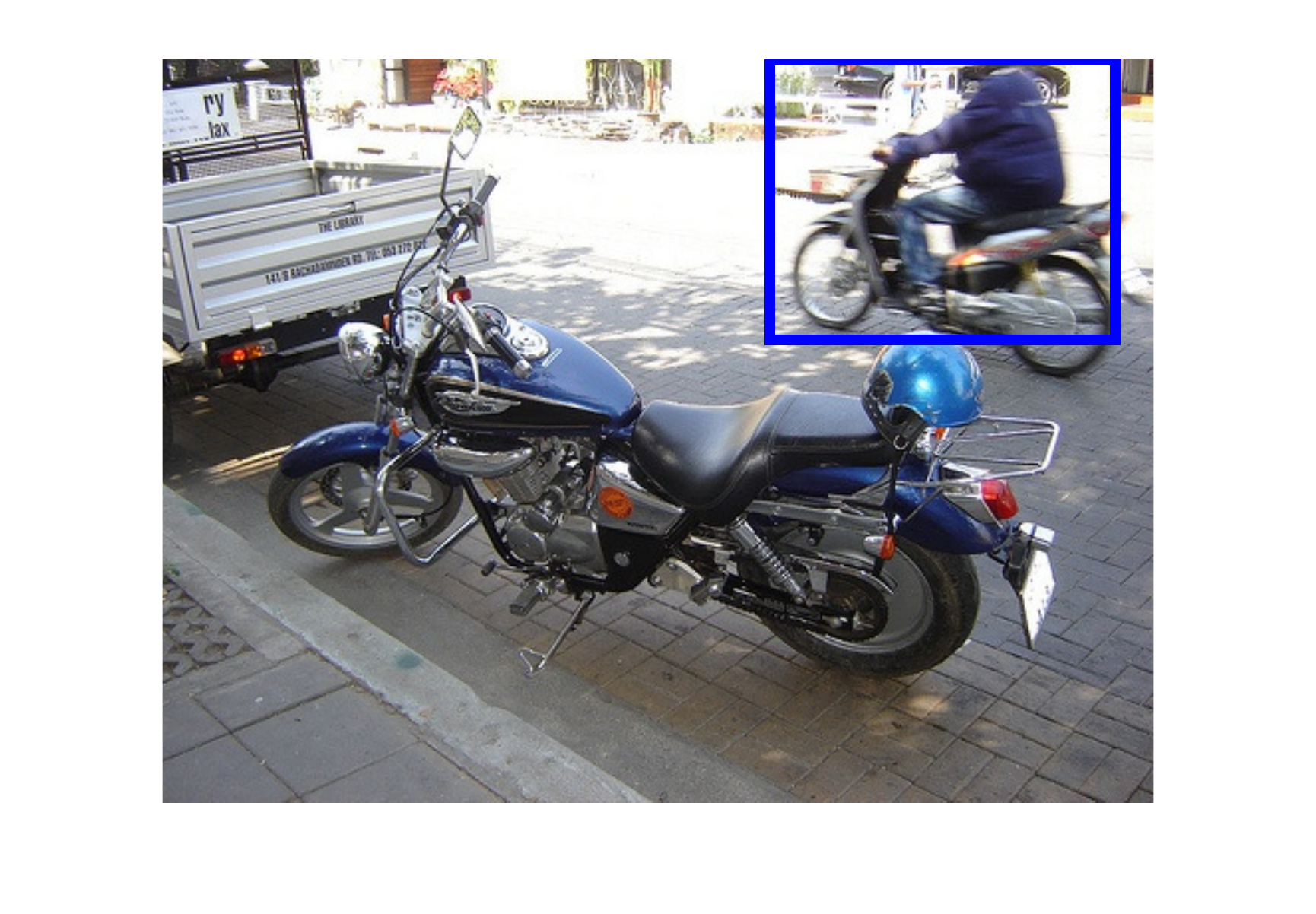}
\caption{person}
\label{fig5d}
\end{subfigure}
\begin{subfigure}{0.2\textwidth}
\includegraphics[width=0.9\linewidth, height=2.7cm]{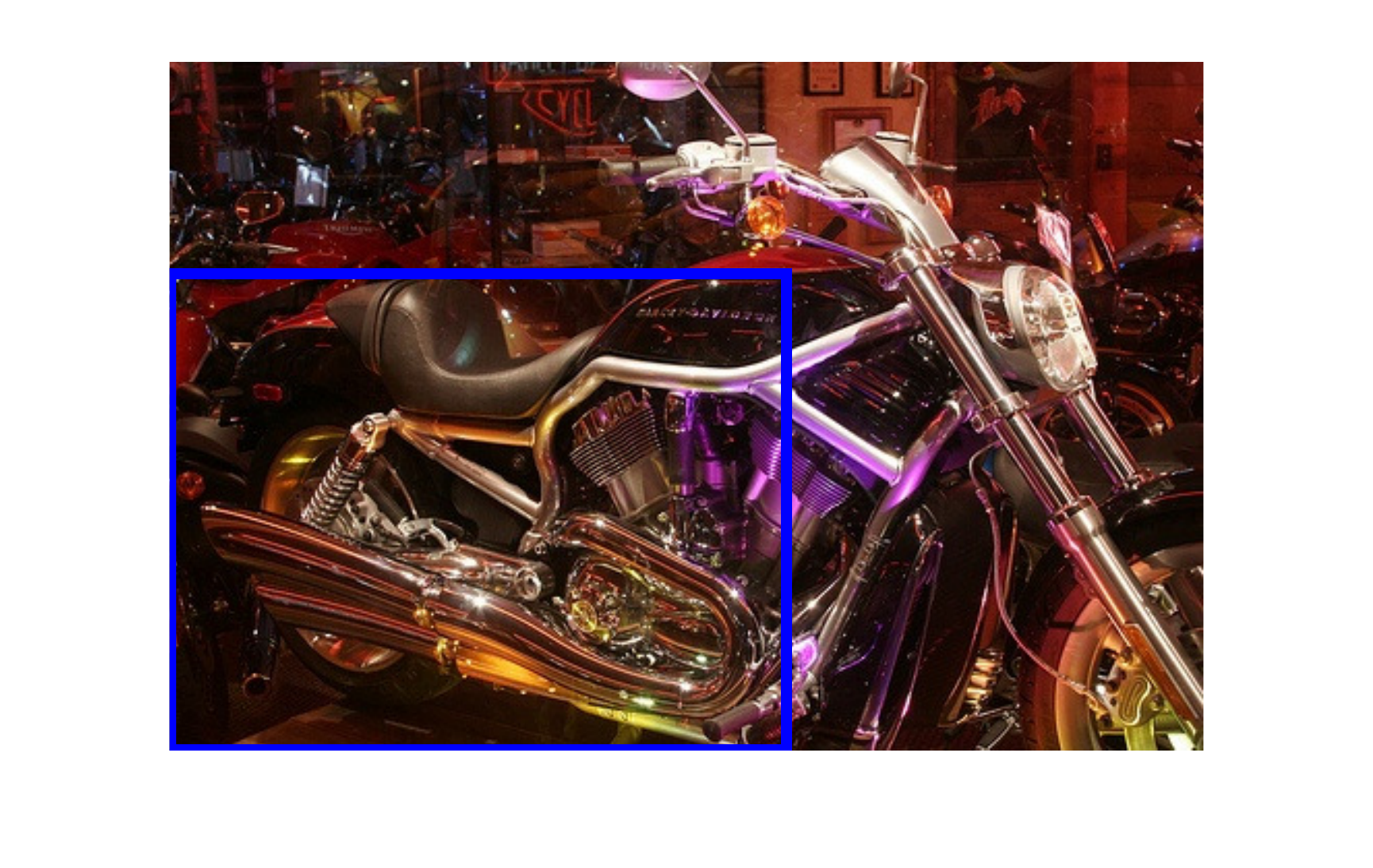}
\caption{bike}
\label{fig5e}
\end{subfigure}
\centering
\begin{subfigure}{0.2\textwidth}
\includegraphics[width=0.9\linewidth, height=2.7cm]{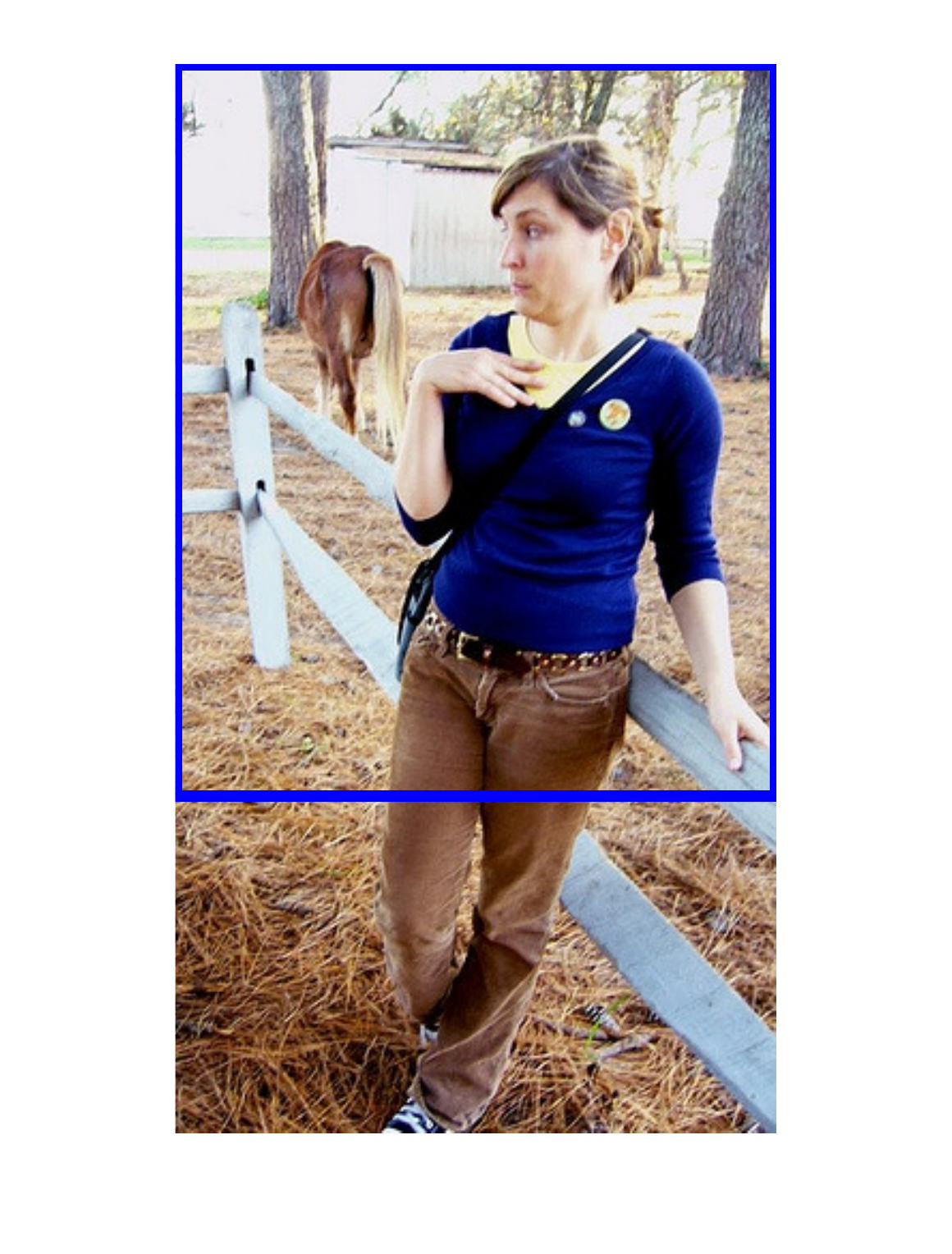}
\caption{horse}
\label{fig5f}
\end{subfigure}
\centering
\begin{subfigure}{0.2\textwidth}
\includegraphics[width=0.9\linewidth, height=2.7cm]{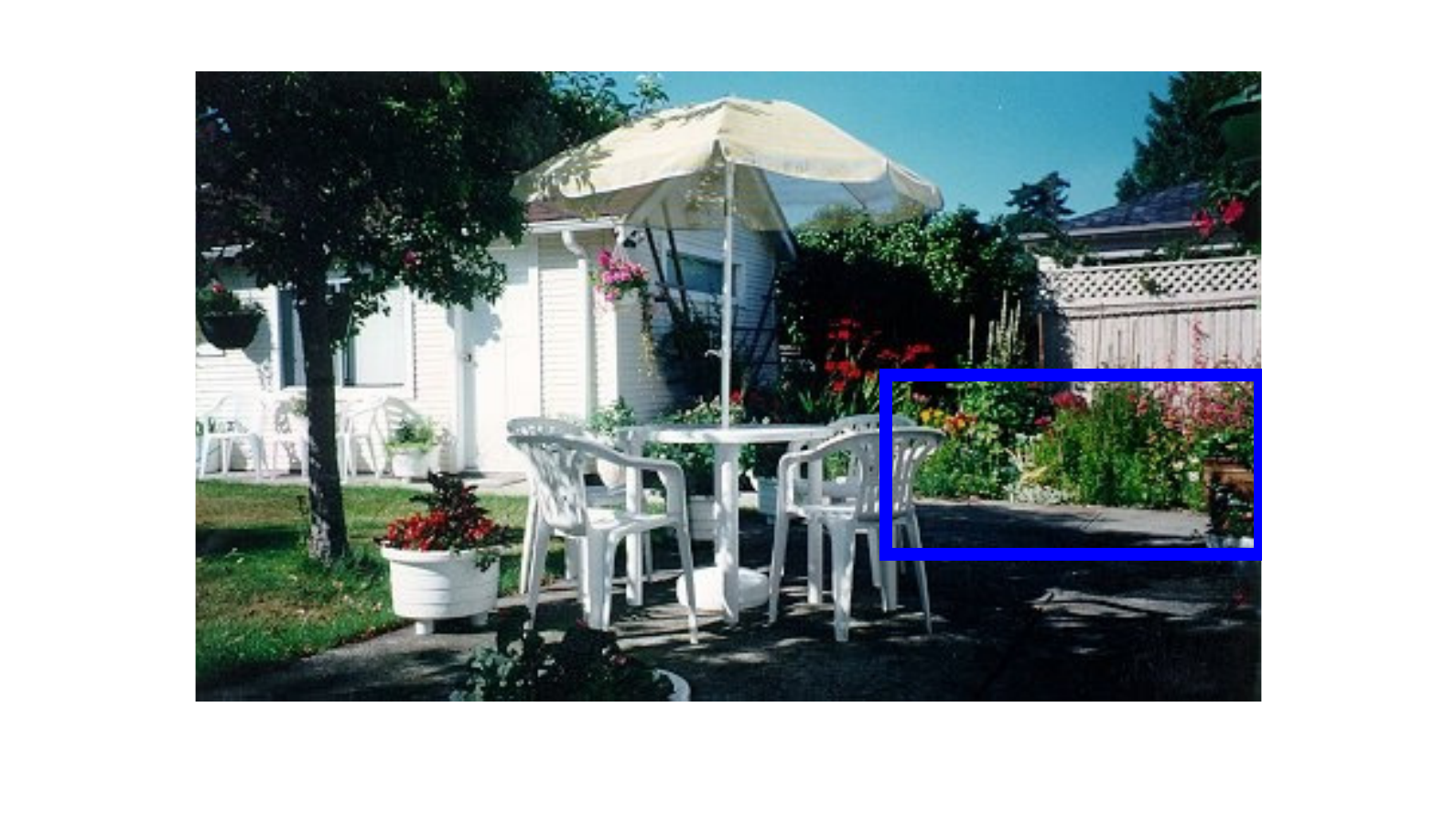}
\caption{plant}
\label{fig5g}
\end{subfigure}
\centering
\begin{subfigure}{0.2\textwidth}
\includegraphics[width=0.9\linewidth, height=2.7cm]{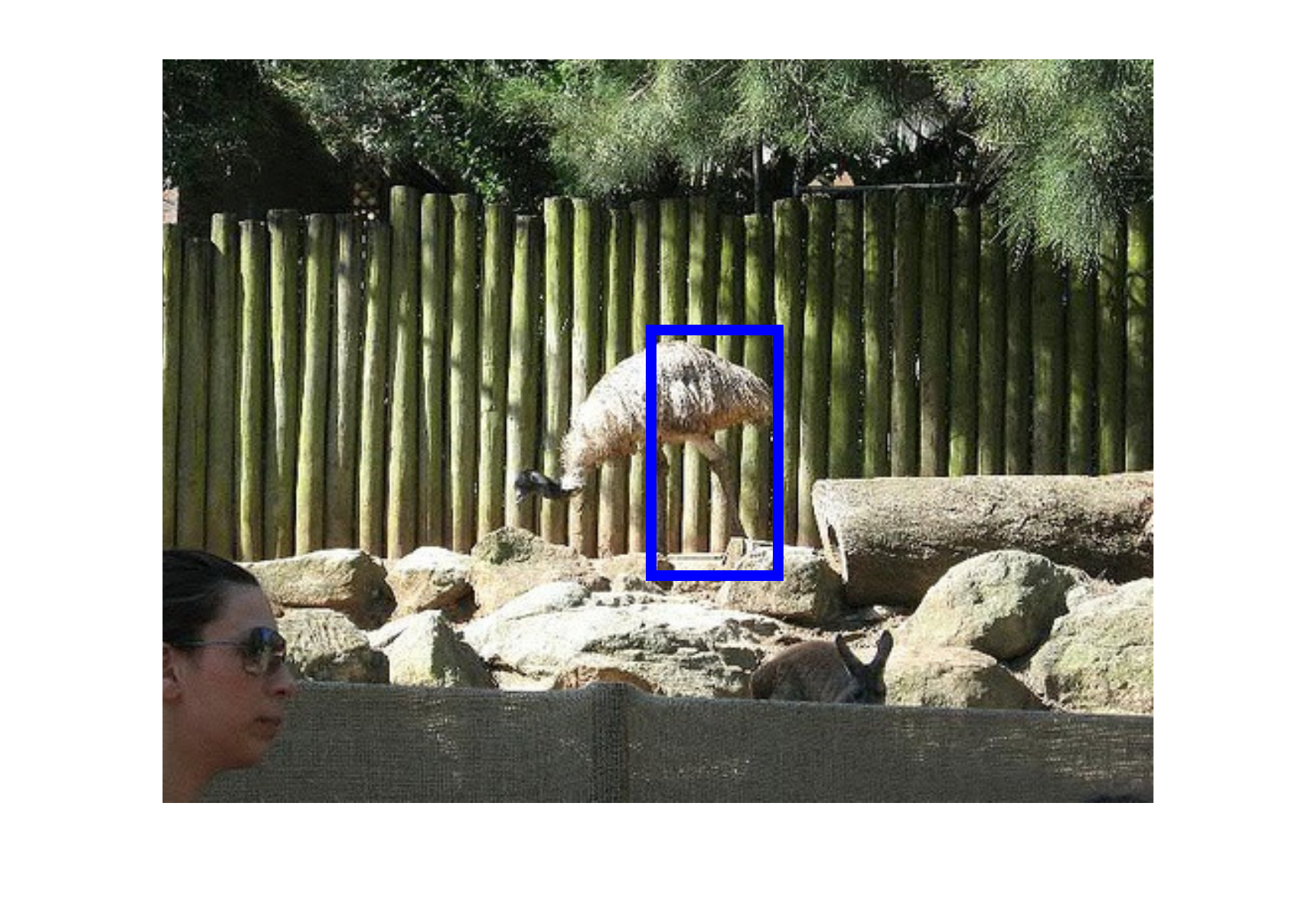}
\caption{person}
\label{fig5h}
\end{subfigure}
\caption{\textbf{Correct (a-d) and incorrect (e-h) detections by APT for some classes}}
\label{fig:detection_results}
\footnotesize{An interesting observation is that some of the incorrect detections (like f) can be corrected by efficient bounding box regression (this is not included in this paper).}
\end{framed}
\end{figure*}

We now consider some visual examples of detections obtained by the APT method. As can be seen from figure~\ref{fig:detection_results}, the method is able to obtain quite good detections. Figures~\ref{fig:detection_results}(a-d) show correct detections as they overlap by more than 50\% with the ground truth boxes. Figures~\ref{fig:detection_results}(e-h) indicate incorrect detection results as the score of these bounding boxes is less than 50\%. For instance in Figure~\ref{fig:detection_results}(f), the bounding box of a horse is quite large and includes the human standing next to the horse. These incorrect detections can be corrected using post-processing bounding box regression approaches that are not adopted in this paper. We further consider image instances in the order in which they are queried for class `motorbike' in figure~\ref{fig:query_order}. We emphasize those that result in a significant rise in accuracy using green border and those that cause a drop using red border.

\section{Conclusion and Future Work}
In this paper, we have proposed principled active learning approaches that outperform the classical active learning methods like entropy and margin sampling for object detection in a weakly supervised setting. Even though naive uncertainty-based active learning methods like simple margin of \cite{Tong:2002:SVM:944790.944793} work well in practice, they suffer from sampling bias \cite{Dasgupta:2011:TFA:1959886.1960197}.

The querying strategies, proposed in this paper, can be used to select influential images for the supervised part of a semi-supervised object detector. This paper also provides the motivation for developing better methods for fixed feature extraction in a weakly supervised setting. For example, models like \cite{DBLP:journals/corr/BilenV15} or their context-aware variants like \cite{kantorov2016} can be used to fine-tune features using only image labels, thereby generating better features.

In future, we would like to incorporate an exploration/exploitation trade-off step in the methods to make them resilient to outliers and noisy annotations \cite{Osugi:2005:BEE:1106326.1106365} or combine several active learning measures like \cite{Baram:2004:OCA:1005332.1005342}, \cite{Pandey:2005:SSA:1066677.1066689}. Clustering based outlier detection methods \cite{5734938} can also be used to filter out possibly occluded annotations during the training step of APT, OPT and MC.

It would also be interesting to look at active learning scenarios where the annotation costs of different images are different \cite{Vijayanarasimhan2011}.

We have used features extracted from Caffe \cite{jia2014caffe}, which used models trained on ILSVRC-2012 dataset. So this paper solves the classic active learning problem for object detection where the features of images (and their respective candidate windows) are given but their annotations need to be actively queried. But \cite{girshick14CVPR} used supervised fine-tuning to generate better features and subsequently improved the object detection results. Such fine-tuning is challenging in case of scarce annotated data and is the next logical step for this paper.

{\small
\bibliographystyle{ieee}
\bibliography{citations}

\begin{thebibliography}{10}\itemsep=-1pt

\bibitem{Baram:2004:OCA:1005332.1005342}
Y.~Baram, R.~El-Yaniv, and K.~Luz.
\newblock Online choice of active learning algorithms.
\newblock {\em J. Mach. Learn. Res.}, 5:255--291, Dec. 2004.

\bibitem{Bay2008346}
H.~Bay, A.~Ess, T.~Tuytelaars, and L.~V. Gool.
\newblock Speeded-up robust features (surf).
\newblock {\em Computer Vision and Image Understanding}, 110(3):346 -- 359,
  2008.
\newblock Similarity Matching in Computer Vision and Multimedia.

\bibitem{Bilen14b}
H.~Bilen, M.~Pedersoli, and T.~Tuytelaars.
\newblock Weakly supervised object detection with posterior regularization.
\newblock In {\em BMVC}, 2014.

\bibitem{DBLP:journals/corr/BilenV15}
H.~Bilen and A.~Vedaldi.
\newblock Weakly supervised deep detection networks.
\newblock {\em CoRR}, abs/1511.02853, 2015.

\bibitem{blaschko2008}
M.~Blaschko and C.~Lampert.
\newblock Learning to localize objects with structured output regression.
\newblock In {\em European Conference on Computer Vision (ECCV)}, pages 2--15,
  Berlin, Germany, Oct. 2008. Max-Planck-Gesellschaft, Springer.

\bibitem{chen14active}
Y.~Chen, H.~Shioi, C.~F. Montesinos, L.~P. Koh, S.~Wich, and A.~Krause.
\newblock Active detection via adaptive submodularity.
\newblock In {\em Proc. International Conference on Machine Learning (ICML)},
  June 2014.

\bibitem{DBLP:journals/corr/CinbisVS15}
R.~G. Cinbis, J.~J. Verbeek, and C.~Schmid.
\newblock Weakly supervised object localization with multi-fold multiple
  instance learning.
\newblock {\em CoRR}, abs/1503.00949, 2015.

\bibitem{Dalal05histogramsof}
N.~Dalal and B.~Triggs.
\newblock Histograms of oriented gradients for human detection.
\newblock In {\em In CVPR}, pages 886--893, 2005.

\bibitem{Dasgupta:2011:TFA:1959886.1960197}
S.~Dasgupta.
\newblock Two faces of active learning.
\newblock {\em Theor. Comput. Sci.}, 412(19):1767--1781, Apr. 2011.

\bibitem{Deselaers2012}
T.~Deselaers, B.~Alexe, and V.~Ferrari.
\newblock Weakly supervised localization and learning with generic knowledge.
\newblock {\em International Journal of Computer Vision}, 100(3):275--293,
  2012.

\bibitem{Pascal}
M.~Everingham, L.~Van~Gool, C.~Williams, J.~Winn, and A.~Zisserman.
\newblock The pascal visual object classes (voc) challenge.
\newblock {\em International Journal of Computer Vision}, 88(2):303--338, 2010.

\bibitem{5255236}
P.~F. Felzenszwalb, R.~B. Girshick, D.~McAllester, and D.~Ramanan.
\newblock Object detection with discriminatively trained part-based models.
\newblock {\em IEEE Transactions on Pattern Analysis and Machine Intelligence},
  32(9):1627--1645, Sept 2010.

\bibitem{girshickICCV15fastrcnn}
R.~Girshick.
\newblock Fast r-cnn.
\newblock In {\em International Conference on Computer Vision ({ICCV})}, 2015.

\bibitem{girshick14CVPR}
R.~Girshick, J.~Donahue, T.~Darrell, and J.~Malik.
\newblock Rich feature hierarchies for accurate object detection and semantic
  segmentation.
\newblock In {\em Computer Vision and Pattern Recognition}, 2014.

\bibitem{He:2004:MVS:1026711.1026715}
J.~He, H.~Tong, M.~Li, H.-J. Zhang, and C.~Zhang.
\newblock Mean version space: A new active learning method for content-based
  image retrieval.
\newblock In {\em Proceedings of the 6th ACM SIGMM International Workshop on
  Multimedia Information Retrieval}, MIR '04, pages 15--22, New York, NY, USA,
  2004. ACM.

\bibitem{4563068}
A.~Holub, P.~Perona, and M.~C. Burl.
\newblock Entropy-based active learning for object recognition.
\newblock In {\em Computer Vision and Pattern Recognition Workshops, 2008.
  CVPRW '08. IEEE Computer Society Conference on}, pages 1--8, June 2008.

\bibitem{jia2014caffe}
Y.~Jia, E.~Shelhamer, J.~Donahue, S.~Karayev, J.~Long, R.~Girshick,
  S.~Guadarrama, and T.~Darrell.
\newblock Caffe: Convolutional architecture for fast feature embedding.
\newblock {\em arXiv preprint arXiv:1408.5093}, 2014.

\bibitem{5206627}
A.~J. Joshi, F.~Porikli, and N.~Papanikolopoulos.
\newblock Multi-class active learning for image classification.
\newblock In {\em Computer Vision and Pattern Recognition, 2009. CVPR 2009.
  IEEE Conference on}, pages 2372--2379, June 2009.

\bibitem{kantorov2016}
V.~Kantorov, M.~Oquab, M.~Cho, and I.~Laptev.
\newblock Contextlocnet: Context-aware deep network models for weakly
  supervised localization.
\newblock In {\em Proc. European Conference on Computer Vision (ECCV), IEEE,
  2016}, 2016.

\bibitem{Kapoor:2010:GPO:1747084.1747105}
A.~Kapoor, K.~Grauman, R.~Urtasun, and T.~Darrell.
\newblock Gaussian processes for object categorization.
\newblock {\em Int. J. Comput. Vision}, 88(2):169--188, June 2010.

\bibitem{Karasev_2014_CVPR}
V.~Karasev, A.~Ravichandran, and S.~Soatto.
\newblock Active frame, location, and detector selection for automated and
  manual video annotation.
\newblock In {\em The IEEE Conference on Computer Vision and Pattern
  Recognition (CVPR)}, June 2014.

\bibitem{Krizhevsky_imagenetclassification}
A.~Krizhevsky, I.~Sutskever, and G.~E. Hinton.
\newblock Imagenet classification with deep convolutional neural networks.
\newblock In {\em Advances in Neural Information Processing Systems}, 2012.

\bibitem{Lampert2009}
C.~H. Lampert and J.~Peters.
\newblock {\em Active Structured Learning for High-Speed Object Detection},
  pages 221--231.
\newblock Springer Berlin Heidelberg, Berlin, Heidelberg, 2009.

\bibitem{Li:2013:AAL:2514950.2516203}
X.~Li and Y.~Guo.
\newblock Adaptive active learning for image classification.
\newblock In {\em Proceedings of the 2013 IEEE Conference on Computer Vision
  and Pattern Recognition}, CVPR '13, pages 859--866, Washington, DC, USA,
  2013. IEEE Computer Society.

\bibitem{DBLP:journals/corr/LiuAESR15}
W.~Liu, D.~Anguelov, D.~Erhan, C.~Szegedy, and S.~E. Reed.
\newblock {SSD:} single shot multibox detector.
\newblock {\em CoRR}, abs/1512.02325, 2015.

\bibitem{Long:2015:MMA:2919332.2920026}
C.~Long and G.~Hua.
\newblock Multi-class multi-annotator active learning with robust gaussian
  process for visual recognition.
\newblock In {\em Proceedings of the 2015 IEEE International Conference on
  Computer Vision (ICCV)}, ICCV '15, pages 2839--2847, Washington, DC, USA,
  2015. IEEE Computer Society.

\bibitem{790410}
D.~G. Lowe.
\newblock Object recognition from local scale-invariant features.
\newblock In {\em Computer Vision, 1999. The Proceedings of the Seventh IEEE
  International Conference on}, volume~2, pages 1150--1157 vol.2, 1999.

\bibitem{mitchell_version_space82}
T.~M. Mitchell.
\newblock Generalization as search.
\newblock {\em Artif. Intell.}, 18(2):203--226, 1982.

\bibitem{Mundy2006}
J.~L. Mundy.
\newblock Object recognition in the geometric era: A retrospective.
\newblock In J.~Ponce, M.~Hebert, C.~Schmid, and A.~Zisserman, editors, {\em
  Toward Category-Level Object Recognition}, pages 3--28. Springer Berlin
  Heidelberg, Berlin, Heidelberg, 2006.

\bibitem{Opelt04}
A.~Opelt, M.~Fussenegger, A.~Pinz, and P.~Auer.
\newblock Generic object recognition with boosting.
\newblock Technical Report TR-EMT-2004-01, {EMT, TU Graz, Austria}, 2004.
\newblock Submitted to the IEEE Transactions on Pattern Analysis and Machine
  Intelligence.

\bibitem{Osugi:2005:BEE:1106326.1106365}
T.~Osugi, D.~Kun, and S.~Scott.
\newblock Balancing exploration and exploitation: A new algorithm for active
  machine learning.
\newblock In {\em Proceedings of the Fifth IEEE International Conference on
  Data Mining}, ICDM '05, pages 330--337, Washington, DC, USA, 2005. IEEE
  Computer Society.

\bibitem{5734938}
R.~Pamula, J.~K. Deka, and S.~Nandi.
\newblock An outlier detection method based on clustering.
\newblock In {\em Emerging Applications of Information Technology (EAIT), 2011
  Second International Conference on}, pages 253--256, Feb 2011.

\bibitem{Pandey:2005:SSA:1066677.1066689}
G.~Pandey, H.~Gupta, and P.~Mitra.
\newblock Stochastic scheduling of active support vector learning algorithms.
\newblock In {\em Proceedings of the 2005 ACM Symposium on Applied Computing},
  SAC '05, pages 38--42, New York, NY, USA, 2005. ACM.

\bibitem{ren15fasterrcnn}
S.~Ren, K.~He, R.~Girshick, and J.~Sun.
\newblock {Faster R-CNN}: Towards real-time object detection with region
  proposal networks.
\newblock {\em arXiv preprint arXiv:1506.01497}, 2015.

\bibitem{Settles10activelearning}
B.~Settles.
\newblock Active learning literature survey.
\newblock Technical report, 2010.

\bibitem{a2ba47b632b84c3f9c744081cf281ff8}
K.~Small and D.~Roth.
\newblock Margin-based active learning for structured predictions.
\newblock {\em International Journal of Machine Learning and Cybernetics},
  1(1-4):3--25, 12 2010.

\bibitem{DBLP:journals/corr/SongGJMHD14}
H.~O. Song, R.~B. Girshick, S.~Jegelka, J.~Mairal, Z.~Harchaoui, and
  T.~Darrell.
\newblock One-bit object detection: On learning to localize objects with
  minimal supervision.
\newblock {\em CoRR}, abs/1403.1024, 2014.

\bibitem{Tong:2002:SVM:944790.944793}
S.~Tong and D.~Koller.
\newblock Support vector machine active learning with applications to text
  classification.
\newblock {\em J. Mach. Learn. Res.}, 2:45--66, Mar. 2002.

\bibitem{Uijlings13}
J.~Uijlings, K.~van~de Sande, T.~Gevers, and A.~Smeulders.
\newblock Selective search for object recognition.
\newblock {\em International Journal of Computer Vision}, 2013.

\bibitem{vedaldi09multiple}
A.~Vedaldi, V.~Gulshan, M.~Varma, and A.~Zisserman.
\newblock Multiple kernels for object detection.
\newblock In {\em Proceedings of the International Conference on Computer
  Vision ({ICCV})}, 2009.

\bibitem{Vijayanarasimhan2011}
S.~Vijayanarasimhan and K.~Grauman.
\newblock Cost-sensitive active visual category learning.
\newblock {\em International Journal of Computer Vision}, 91(1):24--44, 2011.

\bibitem{5995430}
S.~Vijayanarasimhan and K.~Grauman.
\newblock Large-scale live active learning: Training object detectors with
  crawled data and crowds.
\newblock In {\em Computer Vision and Pattern Recognition (CVPR), 2011 IEEE
  Conference on}, pages 1449--1456, June 2011.

\bibitem{Vijayanarasimhan2012}
S.~Vijayanarasimhan and K.~Grauman.
\newblock {\em Active Frame Selection for Label Propagation in Videos}, pages
  496--509.
\newblock Springer Berlin Heidelberg, Berlin, Heidelberg, 2012.

\bibitem{vondrick2011}
C.~Vondrick and D.~Ramanan.
\newblock {Video Annotation and Tracking with Active Learning}.
\newblock In {\em Neural Information Processing Systems (NIPS)}, 2011.

\bibitem{Wang2014}
C.~Wang, W.~Ren, K.~Huang, and T.~Tan.
\newblock Weakly supervised object localization with latent category learning.
\newblock In D.~Fleet, T.~Pajdla, B.~Schiele, and T.~Tuytelaars, editors, {\em
  Computer Vision -- ECCV 2014: 13th European Conference, Zurich, Switzerland,
  September 6-12, 2014, Proceedings, Part VI}, pages 431--445. Springer
  International Publishing, Cham, 2014.

\bibitem{wang2014new}
D.~Wang and Y.~Shang.
\newblock A new active labeling method for deep learning.
\newblock In {\em 2014 International Joint Conference on Neural Networks
  (IJCNN)}, pages 112--119. IEEE, 2014.

\bibitem{Jia12}
J.~Xiao, K.~Ehinger, A.~Oliva, and A.~Torralba.
\newblock Recognizing scene viewpoint using panoramic place representation.
\newblock In {\em Computer Vision and Pattern Recognition (CVPR), 2012 IEEE
  Conference on}, June 2012.

\bibitem{7552533}
Y.~Yan, F.~Nie, W.~Li, C.~Gao, Y.~Yang, and D.~Xu.
\newblock Image classification by cross-media active learning with privileged
  information.
\newblock {\em IEEE Transactions on Multimedia}, PP(99):1--1, 2016.

\bibitem{YangMNCH15}
Y.~Yang, Z.~Ma, F.~Nie, X.~Chang, and A.~G. Hauptmann.
\newblock Multi-class active learning by uncertainty sampling with diversity
  maximization.
\newblock {\em International Journal of Computer Vision}, 113(2):113--127,
  2015.

\bibitem{YGLG12}
A.~Yao, J.~Gall, C.~Leistner, and L.~van Gool.
\newblock Interactive object detection.
\newblock In {\em IEEE Conference on Computer Vision and Pattern Recognition
  (CVPR)}, pages 3242--3249, Providence, RI, USA, 2012. IEEE.

\end{thebibliography}
}

\end{document}